\documentclass[10pt,journal,compsoc]{IEEEtran}
\ifCLASSOPTIONcompsoc
  \usepackage[nocompress]{cite}
\else
  \usepackage{cite}
\fi

\ifCLASSINFOpdf
\else
\fi

\usepackage{amsmath}
\usepackage{amssymb}
\usepackage{algorithmic}
\usepackage{algorithm}
\usepackage{enumerate}
\usepackage{graphicx}
\usepackage{color}
\usepackage{tikz}
\usepackage{colortbl}

\ifCLASSOPTIONcompsoc
  \usepackage[caption=false,font=footnotesize,labelfont=sf,textfont=sf]{subfig}
\else
  \usepackage[caption=false,font=footnotesize]{subfig}
\fi

\usepackage{booktabs}
\usepackage{fixltx2e}
\usepackage{dblfloatfix}
\usepackage{url}
\usepackage{bm}
\usepackage{hyperref}
\usepackage{multirow,tabularx}
\newcolumntype{C}{>{\centering\arraybackslash}X}

\DeclareMathOperator*{\argmin}{\arg\min}

\usepackage{amsthm}

\theoremstyle{plain}
\newtheorem{prop}{Proposition}
\theoremstyle{definition}

\begin{document}
\title{Disorder-invariant Implicit Neural Representation}

\author{Hao Zhu$^\dagger$, Shaowen Xie$^\dagger$, Zhen Liu$^\dagger$, Fengyi Liu, Qi Zhang, You Zhou, Yi Lin, Zhan Ma$^*$, and Xun Cao 
        % <-this % stops a space
\IEEEcompsocitemizethanks{\IEEEcompsocthanksitem The first three authors contributed equally.
\IEEEcompsocthanksitem H. Zhu, S. Xie, Z. Liu, F. liu, Y. Zhou, Z. Ma, X. Cao are with the School of Electronic Science and Engineering, Nanjing University, Nanjing, 210023, China.\protect
E-mail: mazhan@nju.edu.cn
\IEEEcompsocthanksitem Q. Zhang is with the Tencent AI Lab, Shenzhen, 518054, China.
\IEEEcompsocthanksitem Y. Lin is with the Department of Cardiovascular Surgery of Zhongshan Hospital, Fudan University, Shanghai, 200032, China.
%\IEEEcompsocthanksitem This work was supported in part by NSFC under Grants 62101242, 62071219 
}% <-this % stops an unwanted space
\thanks{Manuscript received April 19, 2005; revised August 26, 2015.}
}


\markboth{Journal of \LaTeX\ Class Files,~Vol.~14, No.~8, August~2015}%
{Shell \MakeLowercase{\textit{et al.}}: Bare Demo of IEEEtran.cls for Computer Society Journals}



\IEEEtitleabstractindextext{%
\begin{abstract}
Implicit neural representation (INR) characterizes the attributes of a signal as a function of corresponding coordinates which emerges as a sharp weapon for solving inverse problems. However, the expressive power of INR is limited by the spectral bias in the network training. In this paper, we find that such a  frequency-related problem could be greatly solved by re-arranging the coordinates of the input signal, for which we propose the disorder-invariant implicit neural representation (DINER) by augmenting a hash-table to a traditional INR backbone. Given discrete signals sharing the same histogram of attributes and different arrangement orders, the hash-table could project the coordinates into the same distribution for which the mapped signal can be better modeled using the subsequent INR network, leading to significantly alleviated spectral bias. Furthermore, the expressive power of the DINER is determined by the width of the hash-table. Different width corresponds to different geometrical elements in the attribute space, \textit{e.g.}, 1D curve, 2D curved-plane and 3D curved-volume when the width is set as $1$, $2$ and $3$, respectively. More covered areas of the geometrical elements result in stronger expressive power. Experiments not only reveal the generalization of the DINER for different INR backbones (MLP vs. SIREN)  and  various tasks (image/video representation, phase retrieval, refractive index recovery, and neural radiance field optimization) but also show the superiority over the state-of-the-art algorithms both in quality and speed. \textit{Project page:} \url{https://ezio77.github.io/DINER-website/}
\end{abstract}

}

\maketitle

\IEEEdisplaynontitleabstractindextext

% \fi
\IEEEpeerreviewmaketitle

\IEEEraisesectionheading{\section{Introduction}\label{sec:intro}}

%\IEEEPARstart{I}{NR}~\cite{sitzmann2020implicit} continuously describes a signal, providing the advantages of Nyquist-sampling-free scaling, interpolation, and extrapolation without requiring the storage of additional samples~\cite{martel2021acorn}. 
\IEEEPARstart{I}{NR}~\cite{sitzmann2020implicit}  builds the mapping between the coordinate input and the corresponding attribute of a signal using a neural network, which provides the advantages of Nyquist-sampling-free scaling, interpolation, and extrapolation without requiring the storage of additional samples~\cite{martel2021acorn}. By combining it with differentiable physical mechanisms such as the ray-marching rendering~\cite{mildenhall2020nerf, kellnhofer2021neural}, Fresnel diffraction propagation~\cite{zhu2022dnf} and partial differential equations~\cite{karniadakis2021physics}, INR becomes a universal and sharp weapon for solving inverse problems and has achieved significant progress in various scientific tasks, \textit{e.g.}, the novel view synthesis~\cite{tewari2022advances}, free-hand 3D ultrasound reconstruction~\cite{Shen2023cardiacfield}, intensity diffraction tomography~\cite{liu2022recovery} and multiphysics simulation~\cite{karniadakis2021physics}. %Consequently, it is essential to pursue an INR with shallow and narrow network structure as well as high characterization capabilities to improve the capacity and efficiency for inverse problems.

\begin{figure}[t]
  \centering
  \includegraphics[width=0.8\linewidth]{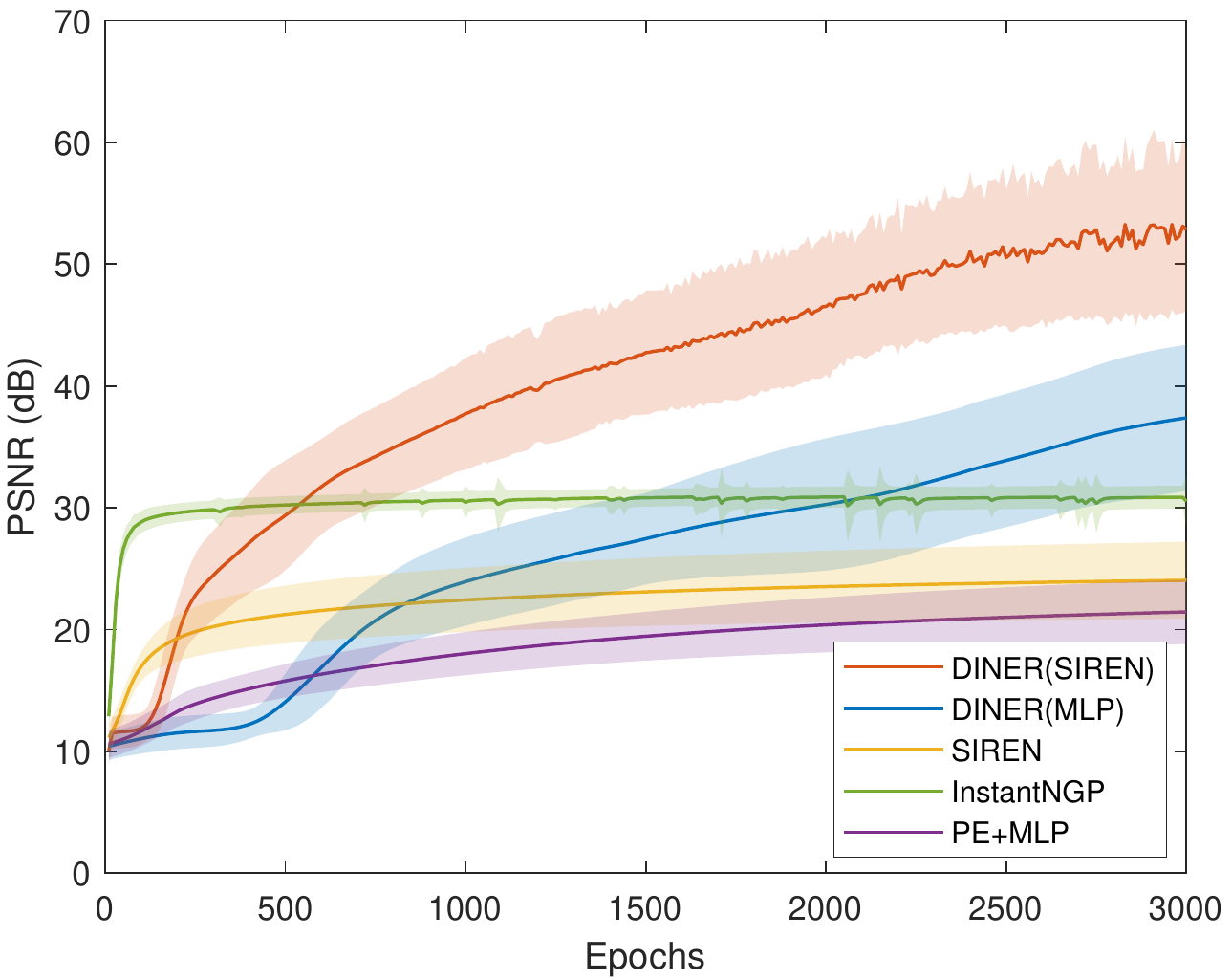}
  % \vspace{-0.2cm}
  \caption{PSNR of various INRs on 2D image fitting over different training epochs.}
  \label{fig:PSNR_over_epochs}
\end{figure}

However, the expressive power of INR is often limited by the underlying network model itself. For example, the spectral bias~\cite{rahaman2019spectral} usually makes the INR easier to represent low-frequency signal components~\cite{tancik2020fourier}. To improve the expressive power of the INR model, previous explorations mainly rely on encoding more frequency bases using either Fourier basis~\cite{mildenhall2020nerf, tancik2020fourier, sitzmann2020implicit, yuce2022structured} or wavelet basis~\cite{fathony2020multiplicative, lindell2021bacon} into the network. However, the length of function expansion is infinite in theory, and a larger model with more frequency bases is running exceptionally slow.

\begin{figure*}[t]
  \centering
  \includegraphics[width=\linewidth]{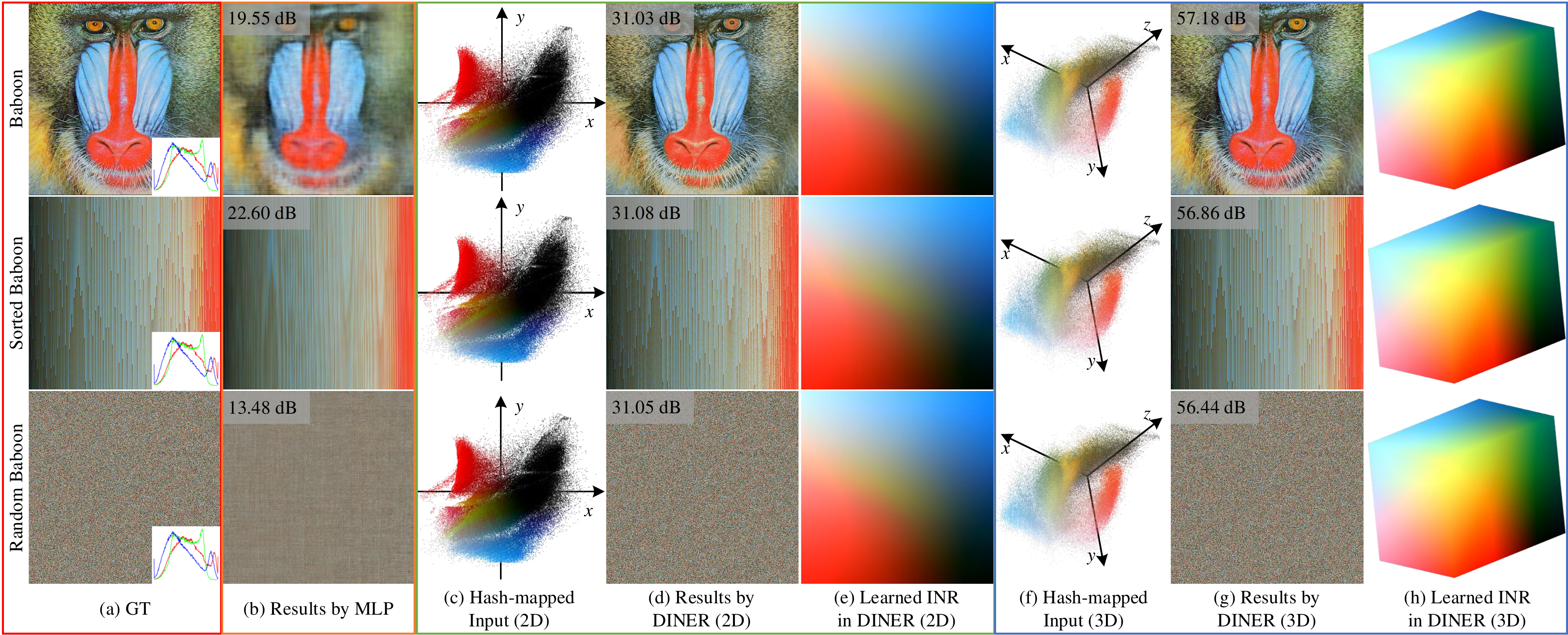}
  \caption{Comparisons of the existing INR and DINER for representing Baboon with different arrangements. From top to bottom, pixels in the Baboon are arranged in different geometric orders, while the histogram is not changed (the right-bottom panel in (a)). From left to right, (a) refers to the ground truth image. (b) contains results by an MLP with positional encoding (PE+MLP)~\cite{tancik2020fourier} at a  size of $2\times 64$. (c) refers to the hash-mapped coordinates with hash-table width $L=2$. (d) refers to the DINER results that use the same-size MLP as (b) and hash-table size $L=2$. (e) refers to the learned INR in DINER results that used coordinates to the trained MLP (hash-table width $L=2$, see Sec.~\ref{sec:method_detail} for more details). (f), (g) and (h) have the same physical meanings as (c), (d), and (h) except using a different hash-table with its width $L=3$.}
  \label{fig:freq_cmp_rgb}
\end{figure*}

 Such a problem is closely related to the frequency spectrum distribution of the input signal. The signal's frequency tells how fast the signal attributes change following the intrinsic order of geometry coordinates. By properly rearranging the order of signal coordinates, we could modulate its frequency spectrum to have more low-frequency components, which makes  the  re-arranged signal better modeled using subsequent INR. Thus, we propose the DINER, in which the input coordinate is first mapped to another index using a hash-table and fed into a classical INR backbone. 
 
 We prove that no matter what geometric orders of the signal attributes are presented initially,  optimizing the hash-table and the network parameters jointly guarantees the same  hash-mapped signal (Fig.~\ref{fig:freq_cmp_rgb}(c) and (f)) that possesses  more low-frequency components. As a result, the expressive power of the existing INR backbones and the task performance are greatly improved. As in Fig.~\ref{fig:PSNR_over_epochs} and Fig.~\ref{fig:freq_cmp_rgb}(d) and (g), a tiny MLP-based INR with shallow and narrow structure can well characterize the input signal with arbitrary arrangement orders under the DINER framework. The use of hash-table trades the storage for fast computation where the caching of its learnable parameters for mapping is usually at a similar size as the input signal, but the computational cost is marginal since the back-propagation for the hash-table derivation is $\mathcal{O}(1)$.

%{\color{red}{We further report that the expressive power of the DINER is could be modeled using parametric functions and the covered areas of these functions in the attribute space determine the power. Different number of input variables of these functions (\textit{i.e.}, the width of the hash-table) corresponds to different geometrical elements, \textit{e.g.}, 1D curve, 2D curved-plane and 3D curved-volume when the width is set as 1, 2 and 3, respectively. As a result, the expressive power of the DINER increases with the increase of the width of the hash-table until it reaches the rank of the signal.
%}}

We further report that the expressive power of the DINER is determined by the width $L$ of the hash-table.   Setting  $L$ at 1, 2, or 3 is equivalent to projecting the  coordinates of the original signal to respective 1D curve, 2D curved plane, and 3D curved volume.  The expressive power of the DINER increases with the increase of the width of the hash-table until it reaches the rank of the signal.
%project the signal elements to  

%could be modeled using parametric functions and the covered areas of these functions in the attribute space determine the power. Different number of input variables of these functions (\textit{i.e.}, the width of the hash-table) corresponds to different geometrical elements, \textit{e.g.}, 1D curve, 2D curved-plane and 3D curved-volume when the width is set as 1, 2 and 3, respectively. As a result, the expressive power of the DINER increases with the increase of the width of the hash-table until it reaches the rank of the signal.

This work is  extended from the preliminary exploration presented in CVPR'23~\cite{xie2023diner}. Compared with the conference version, we model the expressive power of the DINER, and find that the width of the hash-table determines it. To verify this model, additional experiments are conducted on hash-tables with different widths and all experiments in the conference version are re-conducted using the new model for setting the width of hash-table. Apart from this, two additional tasks, \textit{i.e.}, the gigapixel representation and neural radiance field optimization, are conducted.

The main contributions are summarized as follows:
\begin{enumerate}
    % \item {\color{red}{We analyze the influence between the accuracy of INR and the arrangement order of signal.}}
    \item The inferior expressive power of the existing INR model is greatly increased by the proposed DINER, in which a hash-table is augmented to map the coordinates of the original input for better characterization in the succeeding INR model.
    \item The proposed DINER provides consistent mapping and expressive power for signals sharing the same histogram of attributes and different arrangement orders.
    \item The expressive power of the DINER is modeled using parametric functions. The covered areas of the functions determine the power and could be enlarged significantly by increasing the hash-table's width until the signal's rank.
    \item The proposed DINER is generalized in various tasks, including the representation of 2D images and 3D videos, and more vision tasks such as the phase retrieval in lensless imaging, the 3D refractive index recovery in intensity diffraction tomography, as well as neural radiance field optimization, reporting significant performance gains to the existing state-of-the-arts.
\end{enumerate}

\section{Related Work}\label{sec:related_work}

\subsection{INR and inverse problem optimization}

INR (sometimes called the neural field or coordinate neural network) builds the mapping between the coordinate and its signal value using a neural network, promising continuous and memory-efficient modeling for various signals such as the 1D audio~\cite{gao2022objectfolder}, 2D image~\cite{tancik2020fourier}, 3D shape~\cite{park2019deepsdf}, 4D light field~\cite{sitzmann2021light} and 5D radiance field~\cite{mildenhall2020nerf}. Accurate INR for these signals could be supervised directly by comparing the network output with the ground truth, or an indirect way that calculates the loss between the output after differentiable operators and the variant of the ground truth signal. Thus, INR becomes a universal tool for solving inverse problems because the forward processes in these problems are often well-known. INR has been widely applied in the optimization of inverse problems in several disciplines, such as computer vision and graphics~\cite{tewari2022advances}, computational physics~\cite{karniadakis2021physics}, clinical medicine~\cite{Shen2023cardiacfield, yeung2021implicitvol}, biomedical engineering~\cite{zhu2022dnf, liu2022recovery}, material science~\cite{chen2020physics} and fluid mechanics~\cite{raissi2020hidden, reyes2021learning}.

\subsection{Encoding high-frequency components in INR}
Most INRs utilize the MLP as function approximators for characterizing the signal's attributes. According to the approximation theory, an MLP network could approximate any function~\cite{leshno1993multilayer}. However, there is a spectral bias~\cite{rahaman2019spectral} in the network training, resulting in low performance of INR for high-frequency components. Several attempts have been explored to overcome this bias and could be classified into two categories, \textit{i.e.}, the function expansion and the parametric encodings.

The function expansion idea treats the INR fitting as a function approximation using different bases. Mildenhall et al.~\cite{mildenhall2020nerf} encoded the input coordinates using a series of $\sin / \cos$ functions with different frequencies and achieved great success in radiance field representation. The strategy of frequency-predefined $\sin / \cos$ functions is further improved with random Fourier features, which has been proved to be effective in learning high-frequency components both in theory and in practical~\cite{tancik2020fourier}. Landgraf et al.~\cite{landgraf2022pins} alleviated the bias by proposing a progressive positional encoding. Because the dominant lower-frequency content is removed at each level, improved performance is achieved for reconstructing scenes with wide frequency bands. Sitzmann et al.~\cite{sitzmann2020implicit} replaced the classical ReLU activation with periodic activation function (SIREN). The different layers of the SIREN could be viewed as increasing different frequency supports of a signal~\cite{yuce2022structured}. SIREN is suited for representing complex natural signals and their derivatives. Apart from the Fourier expansion, Fathony et al.~\cite{fathony2020multiplicative} represented a complex signal by a linear combination of multiple wavelet functions (MFN), where the high-frequency components could be well modeled by modulating the frequency in the Gabor filter. Lindell et al.~\cite{lindell2021bacon} developed MFN and proposed the band-limited coordinate networks (BACON), where the frequency at each network layer could be specified at initialization. Yang et al.~\cite{yang2022polynomial} generalized the MFN and the BACON as the polynomial neural fields (PNFs). Furthermore, they proposed the Fourier PNFs where different components of the signal could be manipulated and are applied in the texture transfer and scale-space interpolation. These methods have achieved significant advantages in representing high-frequency components compared with the standard MLP network. However, \textit{the performance of these INRs are limited by the frequency distribution of a signal-self, and often require a deeper or wider network architecture to improve the fitting accuracy}.

From the perspective of the parameter encoding~\cite{liu2020neural,takikawa2021neural,chabra2020deep,jiang2020local}, each input coordinate is encoded using learned features which are fed into an MLP for fitting. Takikawa et al.~\cite{takikawa2021neural} divided the 3D space using a sparse voxel octree structure where each point is represented using a learnable feature vector from its eight corners, achieving real-time rendering of high-quality signed distance functions. Martel et al.~\cite{martel2021acorn} divided the coordinate space iteratively during the INR training (ACORN), where the encoding features for each local block are obtained by a coordinate encoder network and are fed into a decoder network to obtain the attribute of a signal. ACORN achieves nearly 40 dB PSNR for fitting gigapixel images for the first time. Muller et al.~\cite{muller2022instant} replaced the coordinate encoder network with a multi-resolution hash-table. Because the multi-resolution hash-table has higher freedom for characterizing coordinates' features, only a tiny network is used to map the features and the attribute values of a signal. Despite the superiority of faster convergence and higher accuracy in parameter encoding, two of the key questions are still not answered, \textit{i.e.}, \textit{what are the geometrical meanings of these features? How many features are sufficient?}

Compared with these methods, the hash-table in the proposed DINER unambiguously projects the input coordinate into another, or in other words, mapping the signal into the one with more low-frequency components. Different hash-table width refers to mapping space with different dimensions, resulting in different expressive power. The expressive power increases with the increase of the hash-tabel until the rank of the signal.  As a result, a tiny network could achieve very high accuracy compared with previous methods.

\section{Performance of INR}
\subsection{Background of the expressive power of INR}
Following the discussion by Yuce et al.~\cite{yuce2022structured}, an INR with a 1D input $x$ could be modeled as a function $f_{\theta}(x)$ that maps the input coordinate $x$ to its attribute, that
\begin{equation}
\begin{aligned}
    \mathbf{z}^{0}&=\gamma (x), \\
    \mathbf{z}^{j}&=\rho^{j}(\mathbf{W}^{j}\mathbf{z}^{j-1}+\mathbf{b}^{j}),\: j=1,...,J-1\\
    f_{\theta}(x)&=\mathbf{W}^{J}\mathbf{z}^{J-1}+\mathbf{b}^{J}
\end{aligned},
\label{eqn:MLP_structure}
\end{equation}
where $\gamma(\cdot)$ is the preprocess function which is often used to encode more frequency bases in the network, $\mathbf{z}^{j}$ is the output of the $j$-th layer in INR, $\rho$ is the activation function, $\mathbf{W}^{j}$ and $\mathbf{b}^j$ are the weight and bias matrix in the $j$-th layer, $J$ is the number of layers in INR, $\theta=\{\mathbf{W}^{j},\mathbf{b}^{j}\}_{1}^{J}$ refers to the set of all training parameters in the network. 

By expanding the non-linear activation function $\rho$ as polynomial activation functions, the signals which could be represented by the INR follows the form
\begin{equation}
\label{eqn:INR_expressivepower}
f(x)=\sum_{\omega'\in\mathcal{H}_{\Omega}}c_{\omega'}\sin (\langle\omega',x\rangle+\phi_{\omega'}),
\end{equation}
where $\mathcal{H}_{\Omega}$ is the frequency set~\cite{yuce2022structured} determined by the frequency selected in the preprocess function $\gamma(\cdot)$, \textit{e.g.}, the Fourier encoding~\cite{tancik2020fourier}, or the $\sin$ activation~\cite{sitzmann2020implicit}. In other words, \textit{the expressive power of INR is restricted to functions that can be represented using a linear combination of certain harmonics of the $\gamma(\cdot)$}~\cite{yuce2022structured}.

\subsection{Arrangement order of a signal determines the expressive power of INR}

According to the expressive power of an INR (Eqn.~\ref{eqn:INR_expressivepower}), a signal could be well learned when the encoded frequencies in the INR are consistent with the signal's frequency distribution. However, there are two problems in applying this conclusion,
\begin{enumerate}
    \item The frequency distribution of a signal could not be known in advance, especially in inverse problems. Thus proper frequencies could not be well set in designing the architecture of an INR.
    \item Due to the spectral bias in network training~\cite{rahaman2019spectral}, the low-frequency components in a signal will be learned first, while the high-frequency components are learned in an extremely slow convergence~\cite{ronen2019convergence, bietti2019inductive, heckel2020compressive, tancik2020fourier}.
\end{enumerate}

We notice that most of the signals recorded or to be inversely solved today are discrete signals. The frequency distribution of a discrete signal could be changed by arranging elements in different orders at the cost of additional storage for the arrangement rule, resulting in different satisfactions of Eqn.~\ref{eqn:INR_expressivepower}. Consequently, the expressive power of an INR for representing a signal changes with different arrangement orders.

Fig.~\ref{fig:freq_cmp_rgb} gives an intuitive demonstration. The Baboon\footnote{Baboon is a self-contained image in the Matlab by MathWorks$^\copyright$.} image is arranged in different orders in Fig.~\ref{fig:freq_cmp_rgb}(a). The original image contains rich low-, intermediate- and high-frequency information. By sorting the Baboon according to the intensities of pixels, the high-frequency information in $y$-axis almost disappeared. We then arrange the pixels using a random order. Currently, the Baboon contains much high-frequency information. Then a PE+MLP ($2\times 64$, \textit{i.e.}, 2 hidden layers and 64 neurons per layer with ReLU activation) network is applied to learn the mapping between the coordinates and intensities of these three images (Fig.~\ref{fig:freq_cmp_rgb}(b)). The fitting results differ significantly. The PE+MLP gets the best performance in the sorted image, which contains the most low-frequency information, while the worst results appear in the random sorted image, which contains the most high-frequency information. In summary:

\begin{prop}
\label{prop_signal_order}
Different arrangements of a signal have different frequency distributions, resulting in different expressive power of INR for representing the signal-self.
\end{prop}

\section{Disorder-invariant INR}
\subsection{Hash-mapping for INR}
\label{sec:method_detail}
Given a paired discrete signal $Y=\{(\vec{x}_i,\vec{y}_i)\}_{i=1}^{N}$, where $\vec{x}_i$ be the $i$-th $d_{in}$-dimensional coordinate, and $\vec{y}_i$ be the corresponding $d_{out}$-dimensional signal attribute. Following the analysis mentioned above, an ideal arrangement rule $M^*:\:\mathbb{R}^{d_{in}}\rightarrow \mathbb{R}^{L}$ should meets the following rule,
\begin{subequations}
\label{eqn:condition_best_mapping}
\begin{align}
      {M}^{*} &= \argmin_{M}\sum_{k=1}^{K_M}|\omega_{k}|      \label{eqn:condition_best_mapping:a}\\
      \Omega_{M} &\subseteq  \mathcal{H}_{\Omega},\:\:
      \Omega_{M} =\{\omega_{k}\}_{k=1}^{K_{M}},  \label{eqn:condition_best_mapping:b}
\end{align}
\end{subequations}
where $\Omega_{M}$ is the set of frequency by mapping the signal following the rule $M$, $\mathcal{H}_{\Omega}$ is the supported frequency set of the INR network (Eqn.~\ref{eqn:INR_expressivepower}), $K_{M}$ is the number of frequency in the arranged signal, $|\cdot|$ returns the absolute value of $\cdot$. Noting that, the mapped coordinate has $L$ dimensions, referring to re-arrange points in a $L$-dimensional space instead of the original $d_{in}$-dimensional space. The signal with this arrangement could be well learned since both the problems of improper frequency setting (Eqn.~\ref{eqn:condition_best_mapping:b}) and the spectral bias (Eqn.~\ref{eqn:condition_best_mapping:a}) are taken into account.

However, this strategy requires prior knowledge of the signal distribution, which is only suitable for the compression task. In contrast, it losses the ability to optimize inverse problems where the signal distribution to be optimized could not be achieved in advance. In this subsection, we detail the proposed DINER to handle this problem.

\begin{figure}[t]
  \centering
  \includegraphics[width=\linewidth]{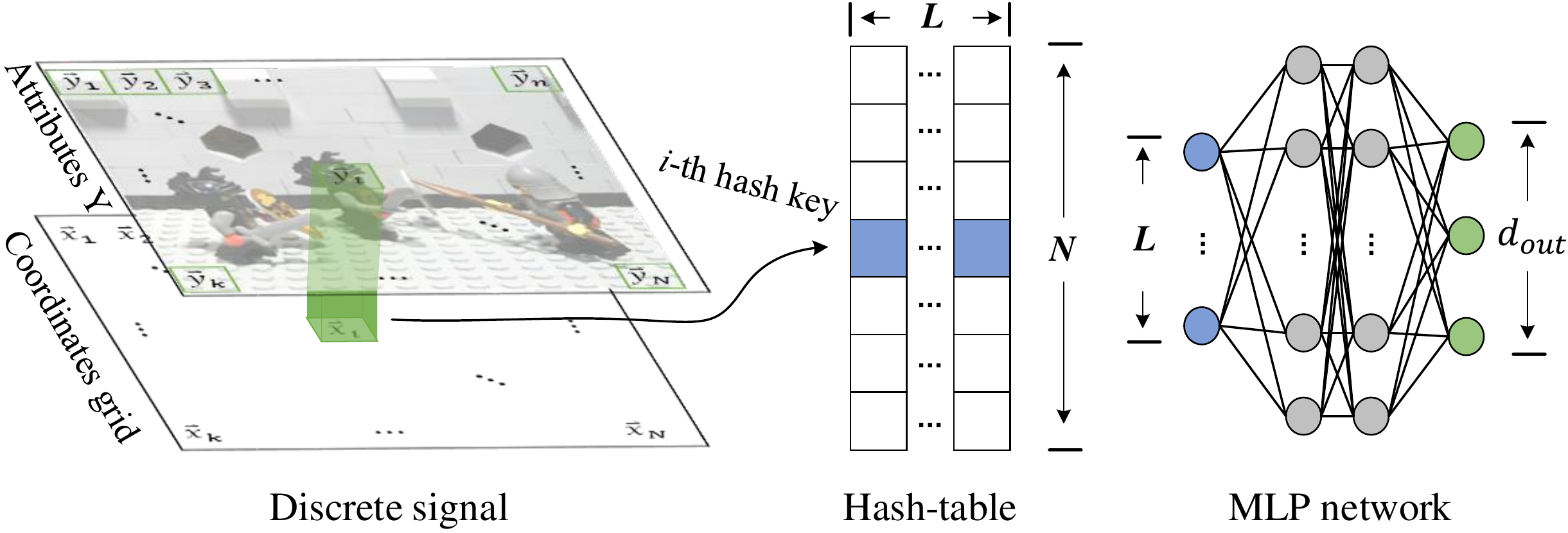}
  \caption{Pipeline of the DINER.}
  \label{fig:hash_table_INR_demo}
  \vspace{-0.3cm}
\end{figure}

We specifically design a full-resolution hash-table $\mathcal{HM}$ to model the mapping $M^*$ in Eqn.~\ref{eqn:condition_best_mapping}. The term `full-resolution' refers to that the hash-table $\mathcal{HM}$ is set as the same length $N$ as the number of elements in $Y$. The width of $\mathcal{HM}$ is set as $L$ ($L$ could be changed for different signals, please refer the Sec.~\ref{sec:diner_expressivepower} for the choice of $L$.). Firstly, the input coordinate $\vec{x}_i$ is used to query the $i$-th hash key $M(\vec{x}_i)$ in the $\mathcal{HM}$. Then the mapped coordinate $M(\vec{x}_i)$ is fed into a standard MLP. All parameters in the $\mathcal{HM}$ are set as learnable, \textit{i.e.}, parameters in $\mathcal{HM}$ and the network parameters will be jointly optimized during the training process. Fig.~\ref{fig:hash_table_INR_demo} demonstrates the above process (The Lego Knight used comes from \cite{stanford_lf_web}.).

Due to the hash-table, the MLP network actually learns the mapped signal. Fig.~\ref{fig:freq_cmp_rgb}(c) and (f) shows the mapped pixels of the Baboon with hash-table width $L=2$ and $3$, respectively. It is noticed that the original grid coordinates (Fig.~\ref{fig:freq_cmp_rgb}(a)) are projected into irregular points (Fig.~\ref{fig:freq_cmp_rgb}(c) and (f)). We sample a mesh evenly according to the minimum and the maximum values in the mapped coordinates, and feed them into the trained MLP, \textit{i.e.}, the Fig.~\ref{fig:freq_cmp_rgb}(e) and (h). For simplicity, the image in Fig.~\ref{fig:freq_cmp_rgb}(f) is later called `learned INR'. The learned INRs differ significantly from the Baboon in that the former are much smoother than the latter and have many low-frequency components (Fig.~\ref{fig:freq_cmp_rgb}(e) and (h)). As a result, high accuracy of MLP fitting could be achieved using the hash-table (Fig.~\ref{fig:freq_cmp_rgb}(d) and (g)).

\subsection{Analysis of disorder-invariance}

When applying the INR to tasks of signal representation and inverse problem optimization, the training or the optimization of the traditional INR and DINER could be modeled as Eqns.~\ref{eqn:loss_func_cmp:CON} and~\ref{eqn:loss_func_cmp:Diner}, respectively.

\begin{subequations}
\label{eqn:loss_func_cmp}
\begin{align}
    \theta^*
    &=\argmin_{\theta}\mathcal{L}\left(\mathcal{P}\left(\{f_{\theta}(\vec{x}_i)\}_{1}^{N}\right), \mathcal{P}\left(\{\vec{y}_i\}_{1}^{N}\right)\right) \label{eqn:loss_func_cmp:CON}\\
    \theta^*,\mathcal{HM}^*
    &=\argmin_{\theta,\mathcal{HM}}\mathcal{L}\left(\mathcal{P}\left(\{f_{\theta}'(\mathcal{HM}(\vec{x}_i))\}_{1}^{N}\right), \mathcal{P}\left(\{\vec{y}_i\}_{1}^{N}\right)\right)\nonumber\\
    &=\argmin_{\theta,\mathcal{HM}}\mathcal{L}\left(\mathcal{P}\left(\{f_{\theta}'(\mathcal{HM}_i)\}_{1}^{N}\right), \mathcal{P}\left(\{\vec{y}_i\}_{1}^{N}\right)\right),\label{eqn:loss_func_cmp:Diner}
\end{align}
\end{subequations}
where $\mathcal{P}$ is a physical process and is an identical transformation for the representation task, $\mathcal{L}$ is the loss function according to the measurements and the reconstructed results, $\mathcal{HM}_i$ is the $i$-th key in the hash-table, {$f_{\theta}'$ has the same parameters setting with $f_{\theta}$ except adding or reducing the number of input variables according to the width $L$ of the hash-table. Because the hash-index operation has no gradient, the first equation in Eqn.~\ref{eqn:loss_func_cmp:Diner} could be simplified to the second one in Eqn.~\ref{eqn:loss_func_cmp:Diner}.

It is noticed that paired relationship between the coordinate $\vec{x}_i$ and value $\vec{y}_i$ in Eqn.~\ref{eqn:loss_func_cmp:CON} is broken in the Eqn.~\ref{eqn:loss_func_cmp:Diner}. There is only one independent variable $\vec{y}_i$ in the loss function (Eqn.~\ref{eqn:loss_func_cmp:Diner}). Assuming parameters in $\theta$ are initialized with the same values, all keys in $\mathcal{HM}$ are set with the same one ($\textit{e.g.}$, 0) in every experiment and the batch size for each iteration is set as $N$, when applying a different order to the signal $Y$, \textit{e.g.}, $Y'=\{\vec{x}_j,\vec{y}_i\}_{j=N,i=1}^{j=1,i=N}$, the training of $\theta$ and $\mathcal{HM}$ for signals $Y$ and $Y'$ share the same optimization progress in every gradient update since all parameters in Eqn.~\ref{eqn:loss_func_cmp:Diner} for $Y$ are equivalent to ones for $Y'$. As a result, the same $\theta^*$ will be optimized while the $\mathcal{HM}'$ of $Y'$ is also an inverse arrangement of the $\mathcal{HM}$ of $Y$. This equivalence is not limited to the $Y'$ with an inverse order; actually, it could be easily proved that the equivalence holds for $Y$ with an arbitrary order. 

Fig.~\ref{fig:freq_cmp_rgb}(c)-(h) illustrate this equivalence. Although the Baboon is arranged with different orders, the hash-table maps them into the same signals (Fig. \ref{fig:freq_cmp_rgb}(c), (e) and \ref{fig:freq_cmp_rgb}(f), (h)), and DINER optimizes them with similar PSNR values ($31.03,\: 31.08,\: 31.05$ when $L=2$ and $57.18,\: 56.86,\: 56.44$ when $L=3$)\footnote{The slight difference in values comes from the floating point errors of GPU for summing matrix with same histogram and different arrangement orders.}. In summary:

\begin{prop}
\label{prop_disorder_invariance}
The DINER is disorder-invariant, and signals with the same histogram distribution of attributes share an optimized network with the same parameter values.
\end{prop}

\subsection{Discussion}
\noindent \textbf{Backbone network.} The backbone of the proposed DINER is not limited to the standard MLP used above; actually, other network structures such as the SIREN could also be integrated with the hash-table and get better performance than the original structure. Please refer to the experimental section for more details.

\noindent \textbf{Complexity.} Although the number of parameters of the hash-table is much larger than the network, the training cost is very small because only one hash-key needs to be updated for training an MLP with batchsize 1. As a result, the computational complexity of the training hash-table is $\mathcal{O}(1)$ in each iteration of training. Note that, the $\mathcal{O}(1)$ complexity holds when the size of hash-table is small. With the increase of the size of the hash-table, more training time are used because the communication cost between the memory and the cache in GPU is increased (see Sec.~6.4 and 6.5 for more details).

\section{Expressive power of the DINER}
\label{sec:diner_expressivepower}
According to the analysis mentioned above, DINER could provide consistent performance for signals with the same histogram distribution of attributes. However, it is still an unknown problem what are the expressive power of the DINER. In this section, we will provide an analysis of this issue.

\begin{figure*}[t]
  \centering
  \includegraphics[width=0.9\linewidth]{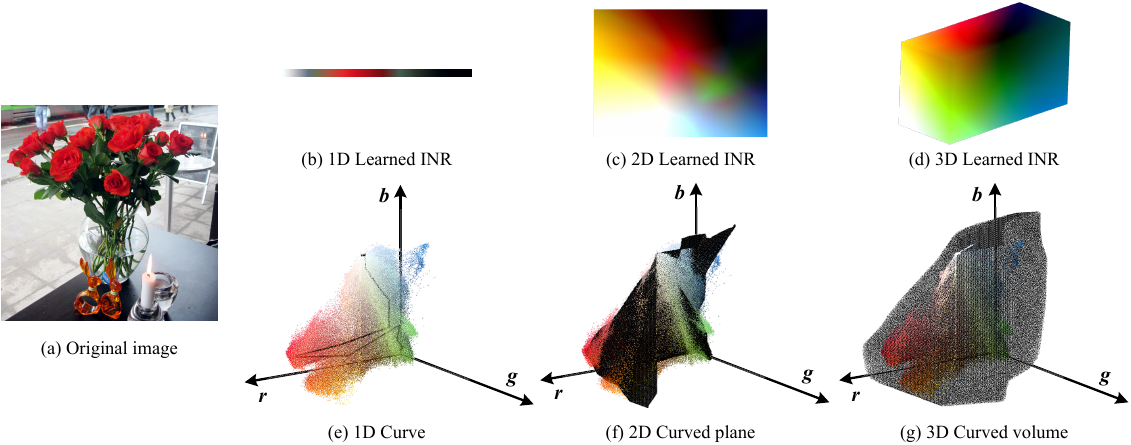}
  \caption{Demonstration of learning a 2D image using DINER with hash-table width $1$, $2$ and $3$, respectively. (a) The original image. (b)-(d) The learned INRs when hash-table has width $1$, $2$ and $3$ respectively. (e)-(g) The corresponding `Hyper-curved-surface' of the learned INRs in the 3D RGB space, where the points in the `Hyper-curved-surface' are labeled with black and the points in the original image are labeled with their original colors. }
  \label{fig:exppower_diner}
  % \vspace{-0.2cm}
\end{figure*}

\subsection{Parametric functions of the DINER}
Revisiting the Eqn.~\ref{eqn:loss_func_cmp:Diner}, the performance of the DINER is independent of the arrangement order of the signal, but is determined by the hash-table $\mathcal{HM}$ and the expressive power of the subsequent network $f_{\theta}'$. Given a hash-table $\mathcal{HM}$, there are three adjustable variables, \textit{i.e.}, the hash-table length $N$, width $L$ and the values in $\mathcal{HM}$. Because the first variable $N$ refers to the number of points in the signal, it could not be changed. Additionally, considering that all values in $\mathcal{HM}$ are learnable, the third variable could not be pre-set before the training process. \textit{As a result, the performance of the DINER is determined by the width $L$ of $\mathcal{HM}$ and the expressive power of the subsequent network $f_{\theta}'$.}

To have a better intuitive experience, the following demonstration and derivation will focus on learning a 2D color image ($d_{out}=3$). According to the expressive power of the INR (Eqn.~\ref{eqn:INR_expressivepower}) and the analysis in Sec.~\ref{sec:method_detail}, the optimization of Eqn.~\ref{eqn:loss_func_cmp:Diner} could be viewed as pursuing learned INRs with the shapes of 1D line, 2D plane and 3D volume when $L=1$, $2$, and $3$, respectively, as well as a hash-table which maps the original coordinate to the one in the 1D line, 2D plane and 3D volume. By drawing these learned INRs in the $d_{out}$-dimensional attribute space according to the attribute of them (\textit{e.g.}, when $d_{out}=3$, a point with attribute $[y_1,y_2,y_3]^{\top}$ will be drawn in the position $[y_1,y_2,y_3]^{\top}$), the 1D line, 2D plane or 3D volume becomes a curved version (they are uniformly called `Hyper-curved-surface' later). Fig.~\ref{fig:exppower_diner} demonstrates the 1/2/3D `Hyper-curved-surface' on learning a 2D RGB image. From Fig.~\ref{fig:exppower_diner}, it is observed that \textit{the problem of the expressive power of the DINER is converted to the issue of how the surface could cover the full points in the attribute space.} 

According to the expressive power of the INR, \textit{i.e.}, the Eqn.~\ref{eqn:INR_expressivepower}, these `Hyper-curved-surface' could be described explicitly using the parametric functions, that
\begin{equation}
\left\{
\begin{aligned}
r&=f_{1}(x)=\sum_{\omega'\in\mathcal{H}_{\Omega}}c_{\omega'}^{1}\sin (\langle\omega',x\rangle+\phi_{\omega'}^{1})\\
g&=f_{2}(x)=\sum_{\omega'\in\mathcal{H}_{\Omega}}c_{\omega'}^{2}\sin (\langle\omega',x\rangle+\phi_{\omega'}^{2})\\
b&=f_{3}(x)=\sum_{\omega'\in\mathcal{H}_{\Omega}}c_{\omega'}^{3}\sin (\langle\omega',x\rangle+\phi_{\omega'}^{3})
\end{aligned}
\right.
\label{eqn:hyper_surface_para_funcs}
\end{equation}
where $x$ is the parametric variable, which refers to the coordinate input to the network or the coordinates in the space of the learned INR. $x$ has the shape $1\times 1$, $2\times 1$ and $3\times 1$ when the width of the $\mathcal{HM}$ is set as $1$, $2$, and $3$, respectively. Meanwhile, the shape of $\omega'$ changes according to the shape of the $x$. From the Eqn.~\ref{eqn:hyper_surface_para_funcs}, the expressive power of the DINER in rgb space is determined by the dimensions of the $x$, or the width of the hash-table and frequency set $\mathcal{H}_{\Omega}$.

\subsection{The width \textit{vs} the frequency}
Supposing all possible `Hyper-curved-surface' that could be represented by Eqn.~\ref{eqn:hyper_surface_para_funcs} is a set $\mathcal{S}_{L}^{\omega}$, where $L$ refers to the width of the hash-table, and $\omega$ is the encoded frequencies in the preprocess function $\gamma(\cdot)$. It could be easily proved that
\begin{equation}
\mathcal{S}_1^{\omega} \subset \mathcal{S}_2^{\omega} \subset \mathcal{S}_3^{\omega}
\label{eqn:hyper_surface_set:1}
\end{equation}
by setting the $2$-nd and the $3$-rd values in the hash-key as $0$ successively (see Appendix A for details). The introduction of the $2$-nd and the $3$-rd variables in the hash-key improves the expressive power of the DINER significantly, because the 2D curved plane could cover much larger space than the 1D curve and the 3D curved-volume increases the space a lot further. On the contrary, encoding more frequencies into the $\gamma(\cdot)$ could only increase the size of the $\mathcal{H}_{\Omega}$ and the space of the Eqn.~\ref{eqn:hyper_surface_para_funcs} linearly.

In summary, the width of the hash-table plays a much important role than adding more frequencies into the preprocess function $\gamma(\cdot)$.

\subsection{On the number of the width}
\label{sec:exp_powers_width}

For representing a 2D color image with $d_{out}=3$, the following formula hold that
\begin{equation}
\forall \:{i,j\geq 3},\:\: \mathcal{S}_i^{0}=\mathcal{S}_j^{0}.
\label{eqn:hyper_surface_set:2}
\end{equation}
To prove it, let's introduce the following parametric functions for representing the rgb color space, 
\begin{equation}
\left\{
\begin{aligned}
r&=\lambda_1+0\lambda_2+0\lambda_3\\
g&=0\lambda_1+\lambda_2+0\lambda_3\\
b&=0\lambda_1+0\lambda_2+\lambda_3
\end{aligned}
\right.,
\label{eqn:color_space_para_funcs}
\end{equation}
where $\lambda = [\lambda_1,\lambda_2,\lambda_3]^{\top}$ is a 3D parameter variable (the space could be represented by the Eqn.~\ref{eqn:color_space_para_funcs} is labeled as $S_{rgb}$ later). It is known that these parametric functions could be easily modeled using a standard MLP (Eqn.~\ref{eqn:MLP_structure}), \textit{e.g.}, constructing a 1 layer MLP with 3 input variables and 3 output values, where no frequency is encoded in the $\gamma(\cdot)$, meanwhile the parameters $\mathbf{W}^{1}$ and $\mathbf{b}^{1}$ are set as
\begin{equation}
\mathbf{W}^{1}=\left[\begin{array}{ccc}
1 & 0 & 0\\
0 & 1 & 0\\
0 & 0 & 1
\end{array}
\right], \:\: 
\mathbf{b}^{1}=\left[\begin{array}{c}
0\\0\\0
\end{array}
\right],
\end{equation}
additionally, no activation function is applied. Because this simple MLP could represent all points in the 3D space and it belongs to the set of possible MLPs constructed by Eqn.~\ref{eqn:hyper_surface_para_funcs} with 3D input $x$ and $\omega=0$, the following formula holds
\begin{equation}
\mathcal{S}_{rgb} = \mathcal{S}_{3}^{0}.
\end{equation}
Because the $\mathcal{S}_3^{0}$ could cover the entire rgb space, it is meaningless to improve the expressive power of the DINER by further increasing the width of the hash-table and the Eqn.~\ref{eqn:hyper_surface_set:2} holds.

Furthermore, the derivation mentioned above is not limited to the rgb color image, actually it could be generalized to other signals with $d_{out}$-dimensional attribute and the Eqns.~\ref{eqn:hyper_surface_set:1}, \ref{eqn:hyper_surface_set:2} are modified as
\begin{equation}
\begin{aligned}
&\mathcal{S}_1^{\omega}\subset\mathcal{S}_2^{\omega}\subset ... \subset\mathcal{S}_{d_{out}}^{\omega}\\
&\forall \:{i\geq d_{out}},\:\: \mathcal{S}_i^{\omega}=\mathcal{S}_{d_{out}}^{\omega}
\end{aligned}.
\label{eqn:color_space_para_funcs:conclusion_lin_independ}
\end{equation}
In summary,  
\begin{prop}
\label{prop_hashtable_width}
The learned INR of the DINER covers the entire attribute space when the width of the hash-table is set as the number of dimensions of the attribute of the signal. Further increment of the width could not improve the expressive power of the DINER. 
\end{prop}

\noindent \textbf{Remark:}\\
The analysis mentioned above focuses on the linearly-independent signals, \textit{i.e.}, any dimension of the attribute could not be linearly represented using other $d_{out}-1$ dimensions. For linearly-dependent signals, it could be easily proved that the parametric function (Eqn.~\ref{eqn:hyper_surface_para_funcs}) could cover the whole attribute space when it has $\rho(Y)$ input variables, where $\rho(Y)$ refers to the rank of the attribute of the signal. The Eqn.~\ref{eqn:color_space_para_funcs:conclusion_lin_independ} is modified as
\begin{equation}
\begin{aligned}
&\mathcal{S}_1^{\omega}\subset\mathcal{S}_2^{\omega}\subset ... \subset\mathcal{S}_{\rho(Y)}^{\omega}\\
&\forall \:{i\geq \rho(Y)},\:\: \mathcal{S}_i^{\omega}=\mathcal{S}_{\rho(Y)}^{\omega}.
\end{aligned}
\label{eqn:color_space_para_funcs:conclusion_lin_depend}
\end{equation}
As a result, 
\begin{prop}
\label{prop_hashtable_width_final}
It is suggested to set width of the hash-table as the rank of the attribute instead of the number of dimensions of the attribute for linearly-dependent signals.
\end{prop}

\begin{figure}[t]
  \centering
  \includegraphics[width=\linewidth]{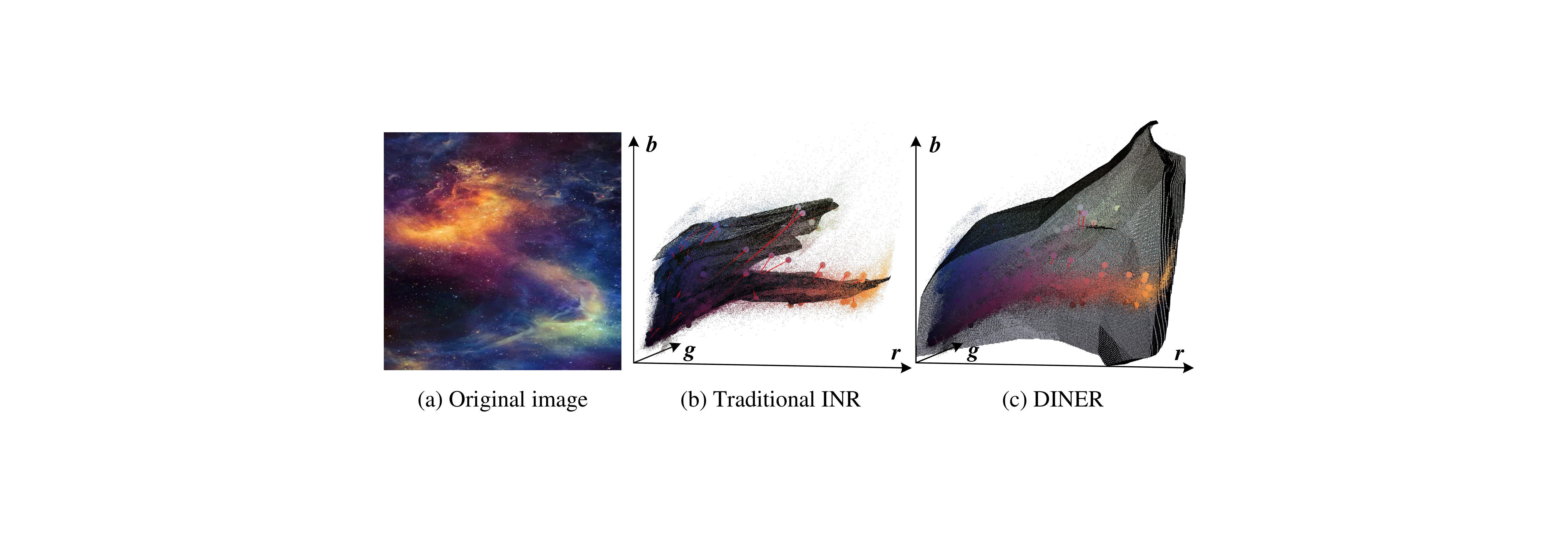}
  \caption{Comparison of the DINER and traditional INR in RGB space. (a) is the original image. The discrete points in (b), (c) are the pixels in (a) where the position is determined by the color. The black curved-planes in (b) and (c) refer to the learned continuous functions of traditional INR and DINER, respectively. Red lines in (b) and (c) refer to the distance between the ground truth and the predicted values (a total of 150 pixel pairs are labelled with larger points for a good visualization.) }
  \label{fig:exppower_diner:cmp_diner_inr}
  % \vspace{-0.2cm}
\end{figure}

\subsection{DINER \textit{vs} Traditional INR in Parametric Space}
Fig.~\ref{fig:exppower_diner:cmp_diner_inr} compares the DINER ($L=2$) and traditional INR in the parametric RGB space using the MLP with same network structure. Limited by the paired relationship between the input coordinate and the attribute, traditional INR focuses on pursuing a complex 2D curved-plane which could pass through all points in the RGB space following the orders of these points in the original image, therefore it is essential to encode complex frequency bases into the parametric functions to support the construction of such a complex curved-plane. On the contrary, the introduction of the hash-table breaks the paired relationship, for this reason the subsequent MLP network in the DINER could focus on finding a curved-plane passing through all points as many as possible and does not take care of the order of connecting them. Once the curved-plane is built, the optimization of hash-table will focus on finding the points in the curved-plane which have minimal distances to the ground truth points. As a result, although traditional INR and DINER share the same parametric function family (Eqn.~\ref{eqn:hyper_surface_para_funcs}) when the same MLP network is used, the introduction of the hash-table makes DINER finding a curved-plane with smaller distances between the ground truth and the predicted one than the traditional INR.

\section{Experiments}
In this section, we will focus on verifying the expressive power of the DINER. All features of the DINER will be verified on the task of 2D image fitting, additionally the tasks of representing gigapixel image and 3D video are used to test the performance of the DINER.

\subsection{Dataset and Algorithm Setup}
For the task of 2D image fitting, 30 high-resolution images with $1200\times 1200$ resolution from the SAMPLING category of the TESTIMAGES dataset~\cite{asuni2014testimages} are used. Each image of the TESTIMAGES dataset is generated using custom Octave/MATLAB software scripts specifically written to guarantee the precise positioning and value of every pixel and contains rich low-, intermediate- and high-frequency information. For the task of representing 3D video, the 'ReadySetGo' and 'ShakeNDry' of the UVG dataset~\cite{mercat2020uvg} are used. 

\begin{table}[t]
  \centering
  \caption{Comparisons of the ratio of frequency distributions between the original image and the learned INR. Two backbones, \textit{i.e.}, the MLP and SIREN, are both compared.}
  \label{tab:freq_dist_w_wo_hash}
  \begin{tabular}{@{}ccccc@{}}
    \toprule
    Freq. bands & \begin{minipage}{0.07\textwidth}
      \includegraphics[width=10mm, height=10mm]{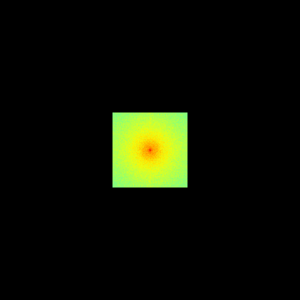}
    \end{minipage} & 
    \begin{minipage}{0.07\textwidth}
      \includegraphics[width=10mm, height=10mm]{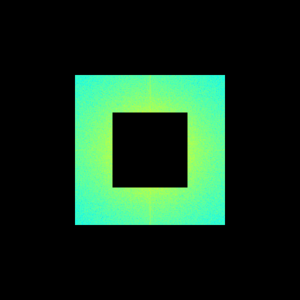}
    \end{minipage} & 
    \begin{minipage}{0.07\textwidth}
      \includegraphics[width=10mm, height=10mm]{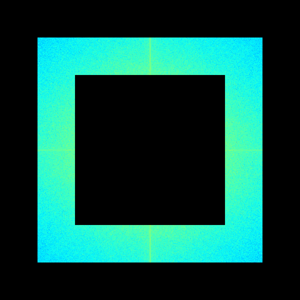}
    \end{minipage} &
    \begin{minipage}{0.07\textwidth}
      \includegraphics[width=10mm, height=10mm]{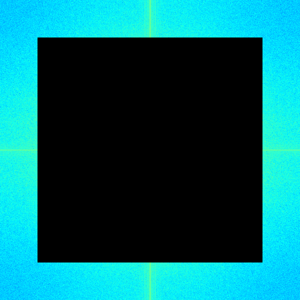}
    \end{minipage}\\
    \midrule
    Original Image & 0.4426 & 0.2484 & 0.1753 & 0.1337 \\
    Learned INR  & \multirow{2}{*}{0.6784} & \multirow{2}{*}{0.1218} & \multirow{2}{*}{0.0973} & \multirow{2}{*}{0.1025} \\
    (MLP) & & & & \\
    Learned INR & \multirow{2}{*}{0.6354} & \multirow{2}{*}{0.1220} & \multirow{2}{*}{0.1149} & \multirow{2}{*}{0.1276} \\
    (SIREN) & & & &\\
    \bottomrule
  \end{tabular}
  % % % \vspace{-0.1cm}
  % \vspace{-0.2cm}
\end{table}

\begin{figure}[t]
  \centering
  \includegraphics[width=\linewidth]{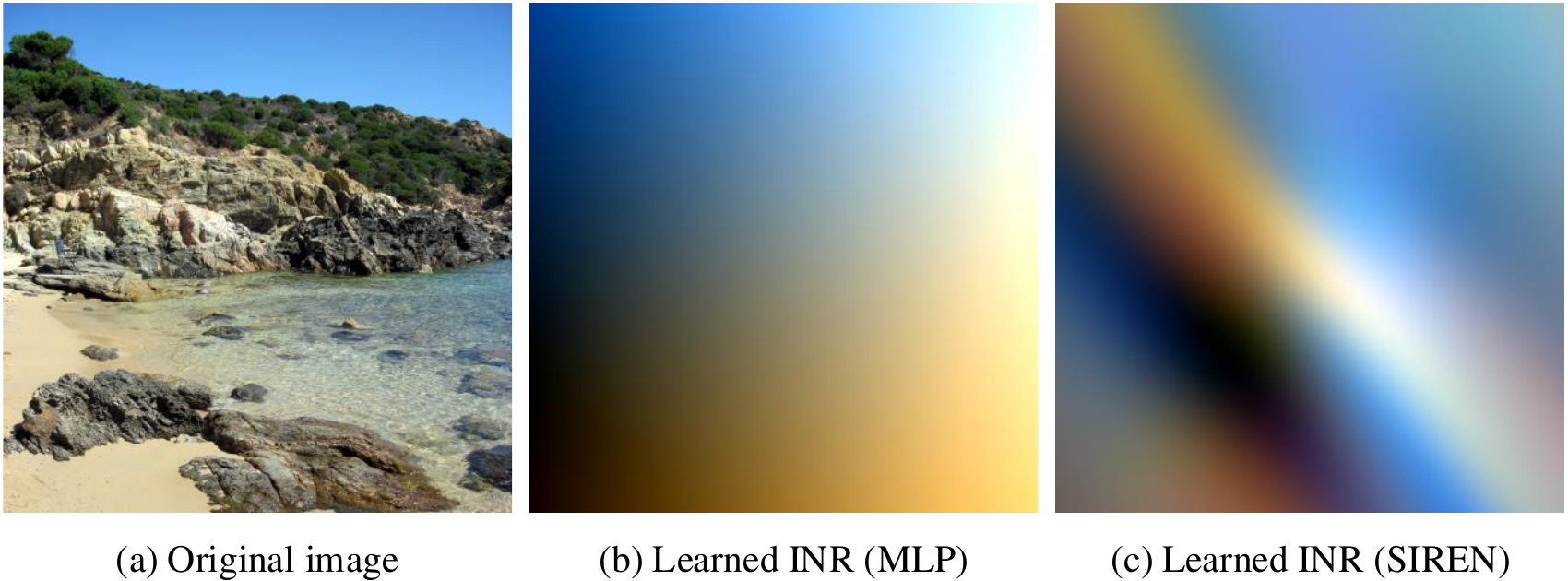}
  \caption{Comparisons of learned INRs in the DINER with MLP and SIREN backbones, respectively.}
  \label{fig:image_cmp_siren_mlp}
  % \vspace{-0.4cm}
\end{figure}

\subsection{Comparisons of frequency distribution with and without hash-table}
We verify the hash-table on both the MLP and SIREN structure with the same network size $2\times 64$ (\textit{i.e.}, 2 layers with 64 neurons per layer). As noticed in Sec.~\ref{sec:method_detail} and Figs.~\ref{fig:freq_cmp_rgb}, the hash-table maps the original signal with more low-frequency contents. Tab.~\ref{tab:freq_dist_w_wo_hash} provides statistics of the mean frequency distributions of the original image and the learned INR over 30 images (the width of the hash-table is set as 2). The ratios of the intermediate- and high-frequency information are all reduced after mapping, while the low-frequency information are increased. 

Compared with the supported frequency set $\mathcal{H}_{\Omega}^{SIREN}$ of the SIREN structure where a default frequency $30$ is used in activation, the $\mathcal{H}_{\Omega}^{MLP}$ of the MLP structure without any frequency encoding contains more low frequencies and less high frequencies. Accordingly, there are more low-frequency information in the mapped image of DINER with MLP backbone than the one with SIREN backbone ($0.6784$ \textit{vs} $0.6354$, Fig.~\ref{fig:image_cmp_siren_mlp}). %Please refer to the supplementary material for more qualitative comparisons. 

\begin{figure*}[!t]
\begin{center}
\centering
\subfloat[$d_{out}=1$]{
	\label{fig:diner_w_dif_width_linear_ind:1}
	\includegraphics[width=0.31\linewidth]{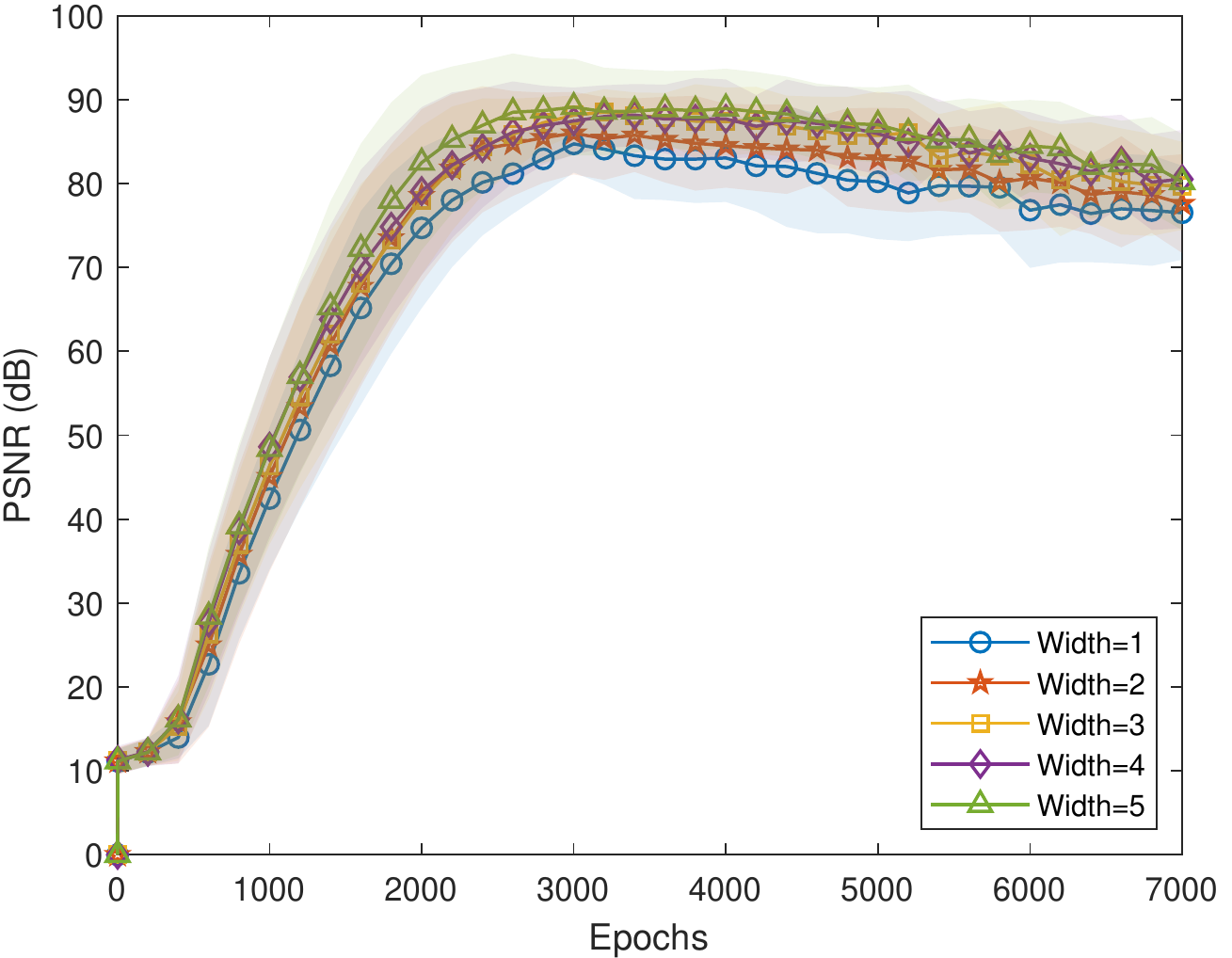}
}
\subfloat[$d_{out}=2$]{
	\label{fig:diner_w_dif_width_linear_ind:2}
	\includegraphics[width=0.31\linewidth]{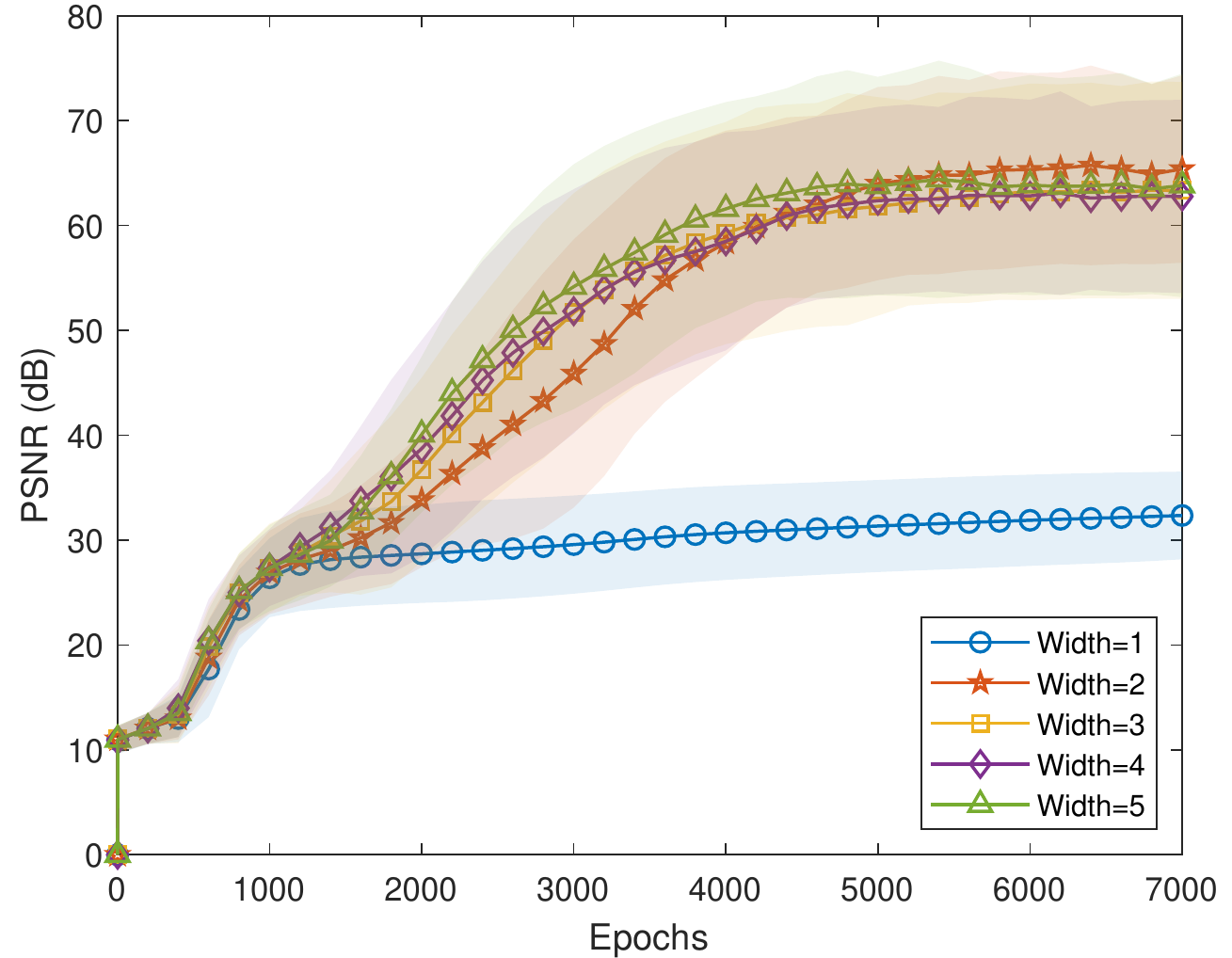}
}
\subfloat[$d_{out}=3$]{
	\label{fig:diner_w_dif_width_linear_ind:3}
	\includegraphics[width=0.31\linewidth]{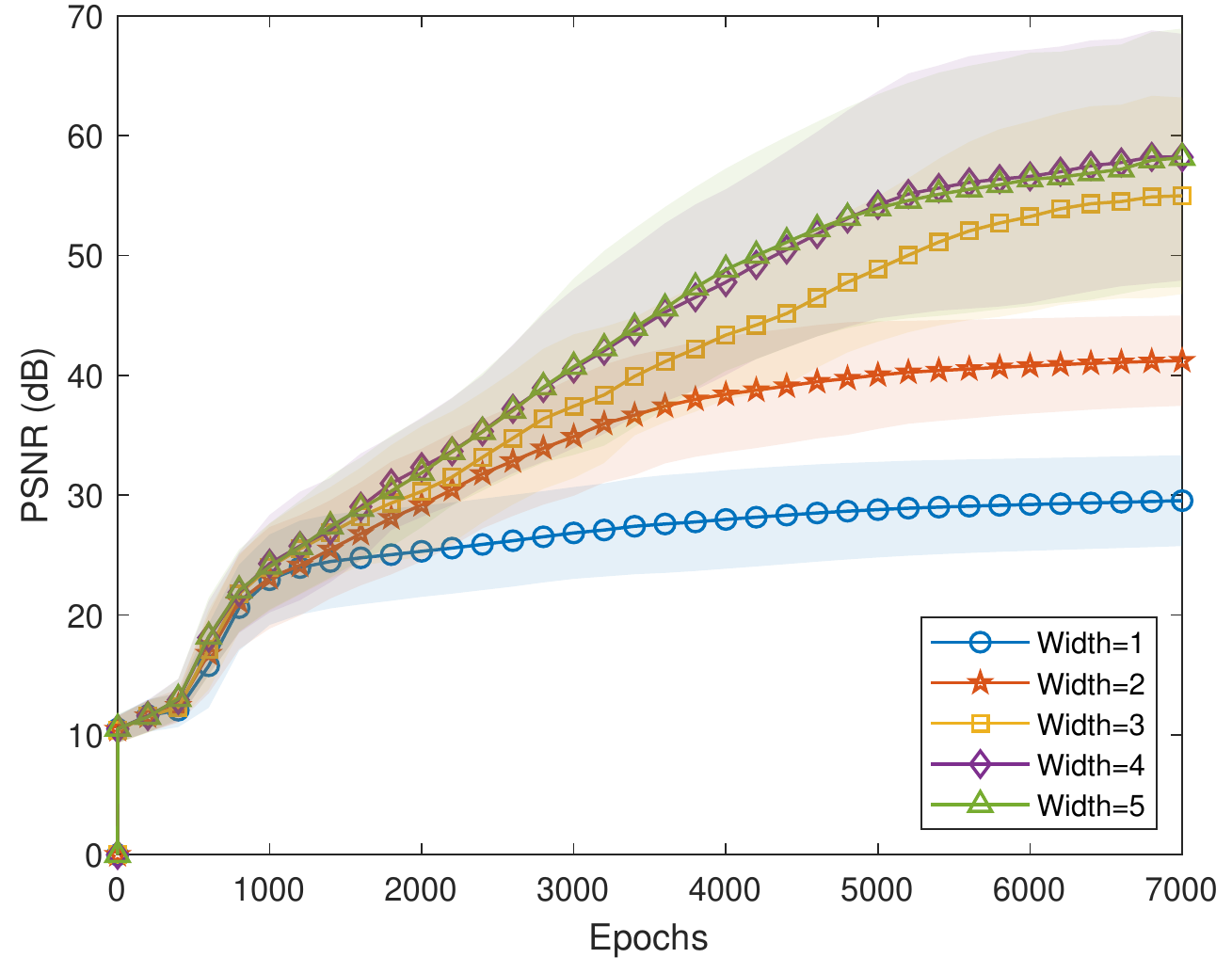}
}
\end{center}
\caption{PSNRs of applying DINER with different widths of hash-table to 2D images with different linearly independent attributes.}
\label{fig:diner_w_dif_width_linear_ind}
\end{figure*}

\begin{figure*}[h]
\begin{center}
\centering
\subfloat[$d_{out}=6$, $\rho(Y)=2$]{
	\label{fig:diner_w_dif_width_linear_dep:1}
	\includegraphics[width=0.31\linewidth]{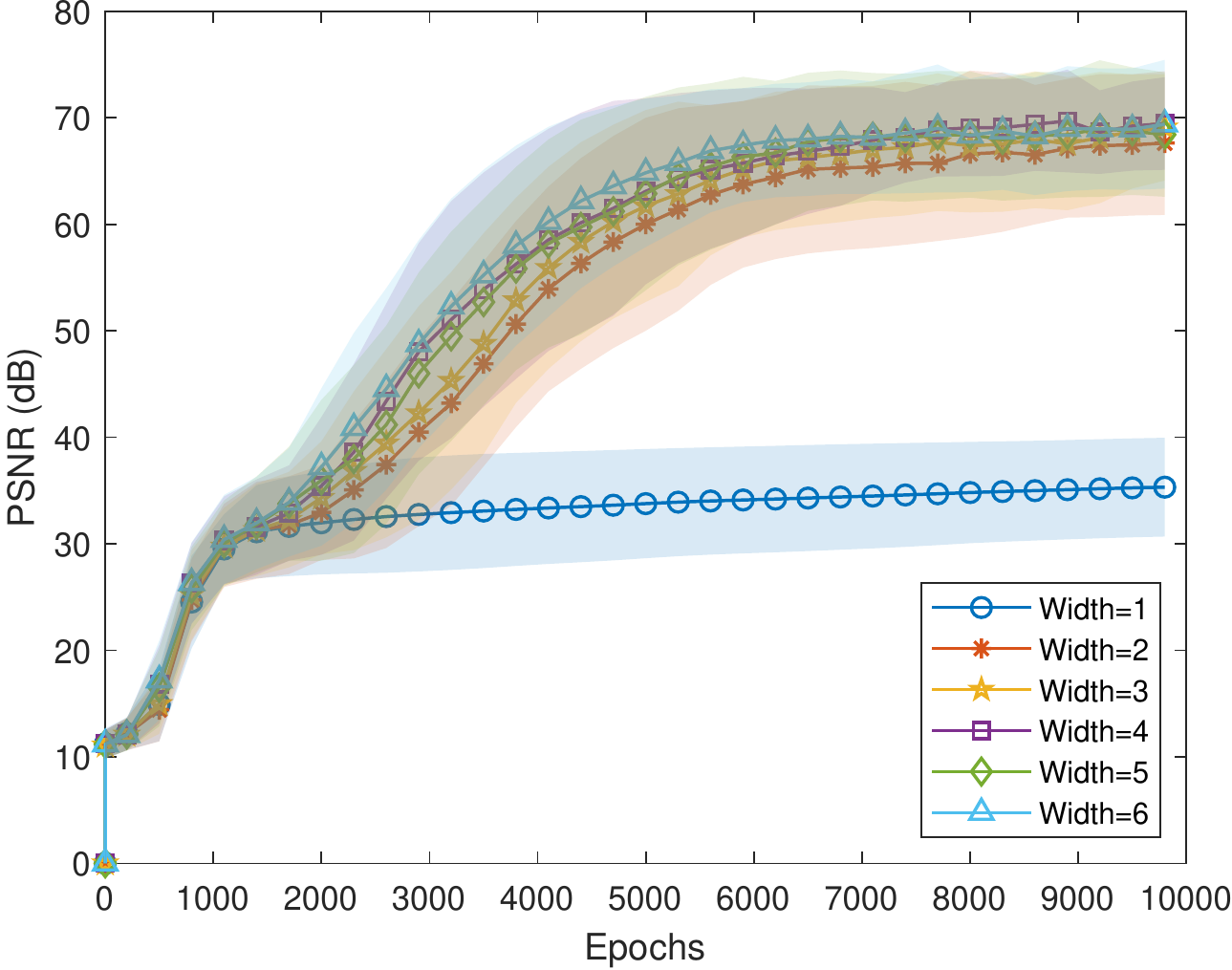}
}
\subfloat[$d_{out}=7$, $\rho(Y)=3$]{
	\label{fig:diner_w_dif_width_linear_dep:2}
	\includegraphics[width=0.31\linewidth]{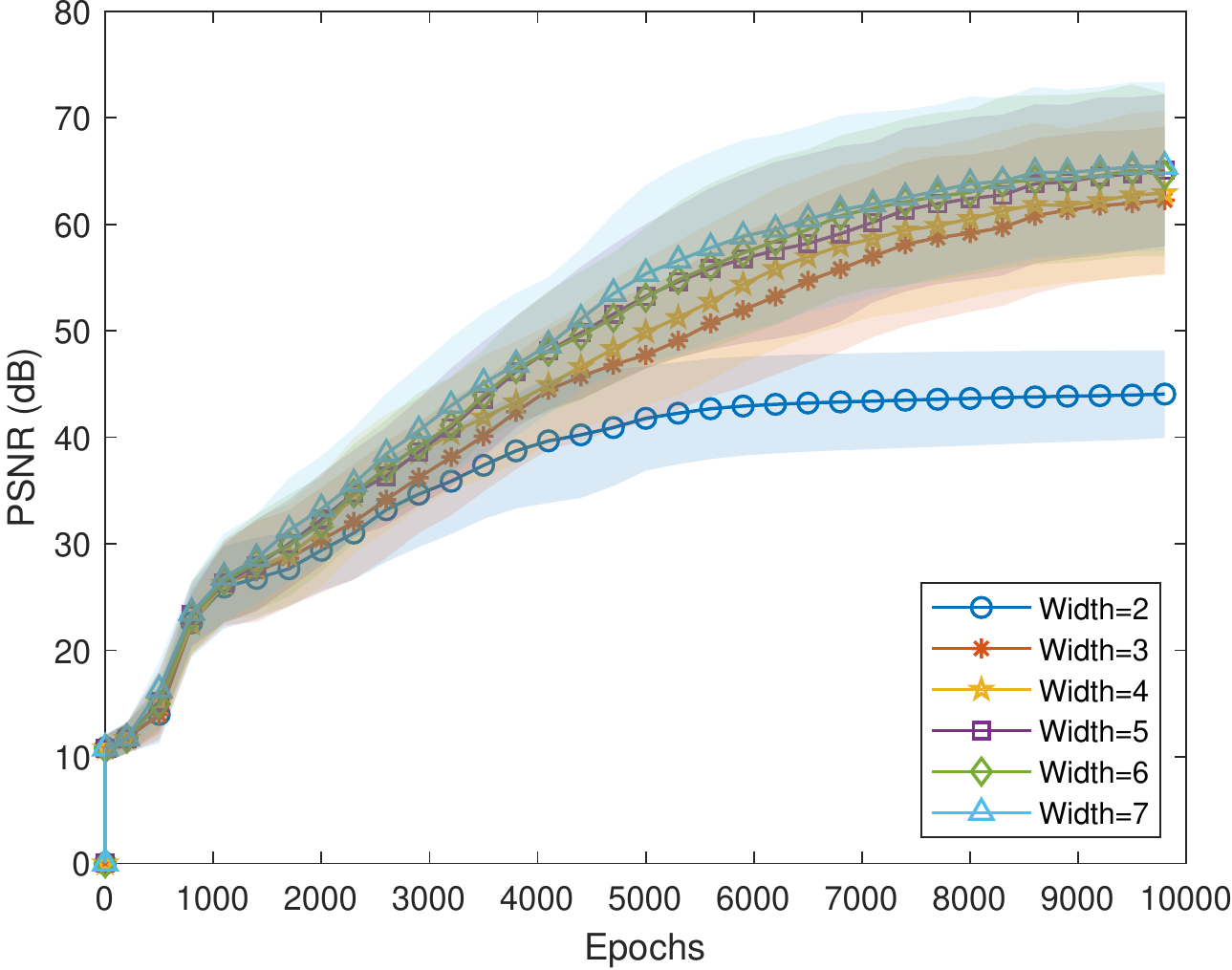}
}
\end{center}
\caption{PSNRs of applying DINER with different widths of hash-table to 2D images with different ranks.}
\label{fig:diner_w_dif_width_linear_dep}
\end{figure*}

\subsection{Comparisons of DINER with different width of hash-table on 2D image fitting}
\label{sec:res_2dimg_dif_width}

In this subsection, we will verify the influence of the width of the hash-table in the DINER towards signals with different number of channels.

\noindent \textbf{Images with linearly independent channels}.
We conduct three experiments to verify the performance of the DINER on images with linearly independent channels, \textit{i.e.}, applying the DINER to the images with 1, 2 and 3-channels. For each type of data, the performance of the DINER is evaluated with different width of the hash-table, \textit{e.g.}, $\{1,2,3,4,5\}$. 

Fig.~\ref{fig:diner_w_dif_width_linear_ind} compares the training curves of different settings. It is noticed that, the PSNR values increase with the increase of the width of the hash-table until the width reaches the number of channels of the images, and they tend to be stable when the width of the hash-table is larger than the number of channels. 

\noindent \textbf{Images with linearly dependent channels}.
We conduct another two experiments to verify the performance of DINER on images with linearly dependent channels. In the first experiment, a dataset including 30 multispectral images with 6 channels are synthesized, where the first 2 channels are copied directly from the $R$ and $G$ channels of the 30 SAMPLING images and the last 4 channels are linearly generated using the first 2 channels, \textit{i.e.}, $\rho(Y)=2$ here. Then, DINER with hash-table width $\{1,2,3,4,5,6\}$ are applied on these images. In the second experiment, a dataset including 30 multispectral images with 7 channels are synthesized, where the first 3 channels are maintained from the original RGB channels and the last 4 channels are also linearly generated from the first 3 channels, \textit{i.e.}, $\rho(Y)=3$ here. DINER with hash-table width $\{2,3,4,5,6,7\}$ are applied on these 7-channels multispectral images.

Fig.~\ref{fig:diner_w_dif_width_linear_dep}(a) and (b) show the training curves of these two experiments, respectively. The PSNR values increase with the increase of the hash-table width until it reaches the ranks of the attributes of these two datasets, respectively.

\begin{figure*}[t]
  \centering
  \includegraphics[width=0.9\linewidth]{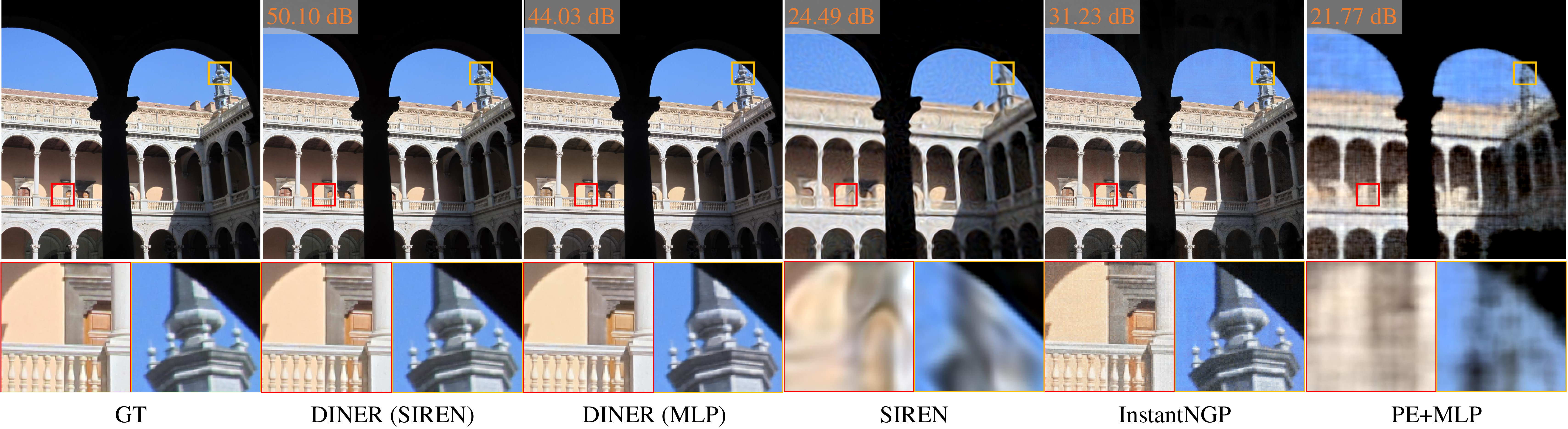}
  \caption{Qualitative comparisons of various methods on 2D image fitting after 3000 epochs. It can be seen that our method provides more clear details compared to SIREN and PE+MLP, especially in the high-frequency boundaries of the bell tower (\textcolor{yellow}{yellow box}). Although InstantNGP achieves better results, its zoom-in results still have many noises due to its linear interpolation during training, as shown in the wall of \textcolor{red}{red box} and the sky of \textcolor{yellow}{yellow box}. In contrast, our method achieves the best results which demonstrate the performance of our DINER. }
  \label{fig:image_res_cmp}
  % \vspace{-0.2cm}
\end{figure*}

\begin{figure*}[t]
  \centering
  \includegraphics[width=0.9\linewidth]{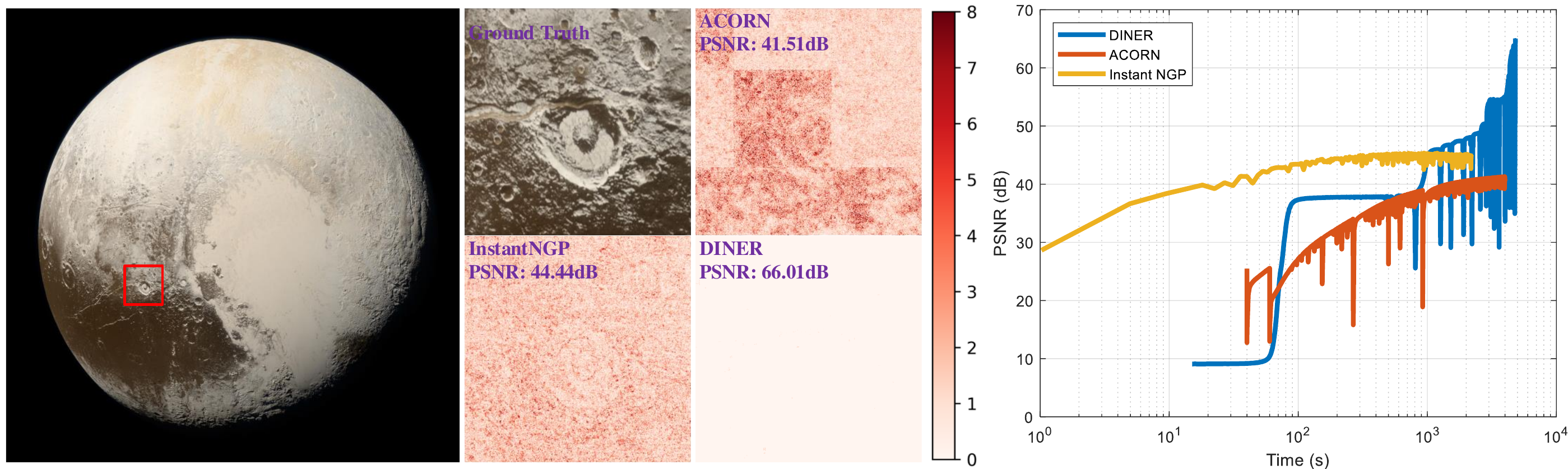}
  \caption{Comparisons of various methods on representing gigapixel image `Pluto'. From left to right, the ground truth, comparisons of error maps of different methods, PSNR curves over training time. It can be seen that the proposed DINER outperforms other methods on gigapixel image representation, particularly in the high-frequency regions. Besides, ACRON has visible block artifacts, while InstantNGP yields smooth results. However, our method performs better than both of these baselines. In addition, the proposed method exhibits better expressive power compared to the baselines.}
  \label{fig:gigapixel_res_cmp}
 % \vspace{-0.3cm}
\end{figure*}

\subsection{Comparisons with the SOTAs on 2D image fitting}
We compare the proposed DINER with the Fourier feature positional encoding (PE+MLP)~\cite{tancik2020fourier}, SIREN~\cite{sitzmann2020implicit} and InstantNGP~\cite{muller2022instant}. Noting that, two backbones, \textit{i.e.}, the standard MLP with ReLU activation and the SIREN with periodic function activation, are all combined with the proposed hash-table to better evaluate the performance. We control the size of the hash-table used in the InstantNGP to guarantee the similar parameters with ours, \textit{e.g.}, $2^{21}$ in the 2D image fitting task while ours has a length of $1200^2 < 2^{21}$. The width of the hash-table in the proposed method and the instantNGP are both set as 3, which is equal to the number of attributes of the 2D image. Apart from this, all 5 methods are trained with the same $L_{2}$ loss between the predicted value and the ground truth, and other parameters are set with the default values by authors.

Fig.~\ref{fig:PSNR_over_epochs} shows the PSNR of various methods at different epochs \footnote{Although the DINER-based methods do not converge to the optimal solution in Fig.~\ref{fig:PSNR_over_epochs}, significant advantages have been achieved over others. As a result, the training curves of more epochs are not plotted, please refer to the yellow curve in Fig.~\ref{fig:diner_w_dif_width_linear_ind}(c) for more details.}. It is noticed that the SIREN and PE+MLP convergences quickly at the early stage and reaches about $24$dB and $21$dB finally. On the contrary, the proposed two methods both provide higher accuracy than backbones. The PSNRs of two backbones for image fitting are increased $30$dB and $17$dB using the hash-table, respectively. Additionally, although the InstantNGP converges very fast to about $30$dB at about 200 epochs, the curve tends to be stable at the last 2800 epochs. The proposed DINER with MLP backbone achieves an advantage of $6$dB than the InstantNGP.

Fig.~\ref{fig:image_res_cmp} shows the qualitative results at 3000 epochs. The proposed methods outperform the SIREN and PE+MLP. Our methods provide more clear details especially in high-frequency boundaries, such as the bell tower (yellow box) in Fig.~\ref{fig:image_res_cmp}. The fitted image of the InstantNGP is very similar to the GT at first sight, however many noises appear in the zoom-in results, for example there are a lot of noisy points in the wall (red box) and the sky (yellow box) of Fig.~\ref{fig:image_res_cmp}, results in a lower PSNR metric. According to the analysis in Sec.~\ref{sec:diner_expressivepower} and Eqn.~\ref{eqn:hyper_surface_set:1}, the DINER(SIREN) is slightly stronger than the DINER(MLP) due to the encoded high frequencies in the periodic activation, as a result, the DINER(SIREN) outperforms DINER(MLP) in the first $3000$ epochs. After $10000$ epochs, the performance of these two methods tends to be similar ($59.64$dB \textit{vs} $59.53$dB, these results are not plotted since they do not affect the conclusion here).

Tab.~\ref{tab:time_image} lists the training time of 5 methods. The InstantNGP is implemented with the tiny-cuda-nn~\cite{muller2022tinycudann}, while other 4 methods are implemented with the Pytorch. All 5 methods are trained on a NVIDIA A100 40GB GPU. The optimization of hash-table requires additional 3 seconds on the SIREN architecture and reduces 20 seconds compared with the classical PE+MLP architecture, verifying the low complexity of optimizing hash-table.

\begin{table}
  \centering
  \caption{Comparisons of training a 2D image with 3000 epochs.}
  \label{tab:time_image}
  \begin{tabular}{@{}lccccc@{}}
    \toprule
     & DINER & DINER & \multirow{2}{*}{SIREN} & \multirow{2}{*}{InstantNGP} & PE \\
     & (SIREN) & (MLP) & & &+MLP\\
    \midrule
    Time& 81.1s & 59.1s & 77.6s & 38s &78.8s \\
    \bottomrule
  \end{tabular}
\end{table}

\begin{figure}[t]
  \centering
  \includegraphics[width=\linewidth]{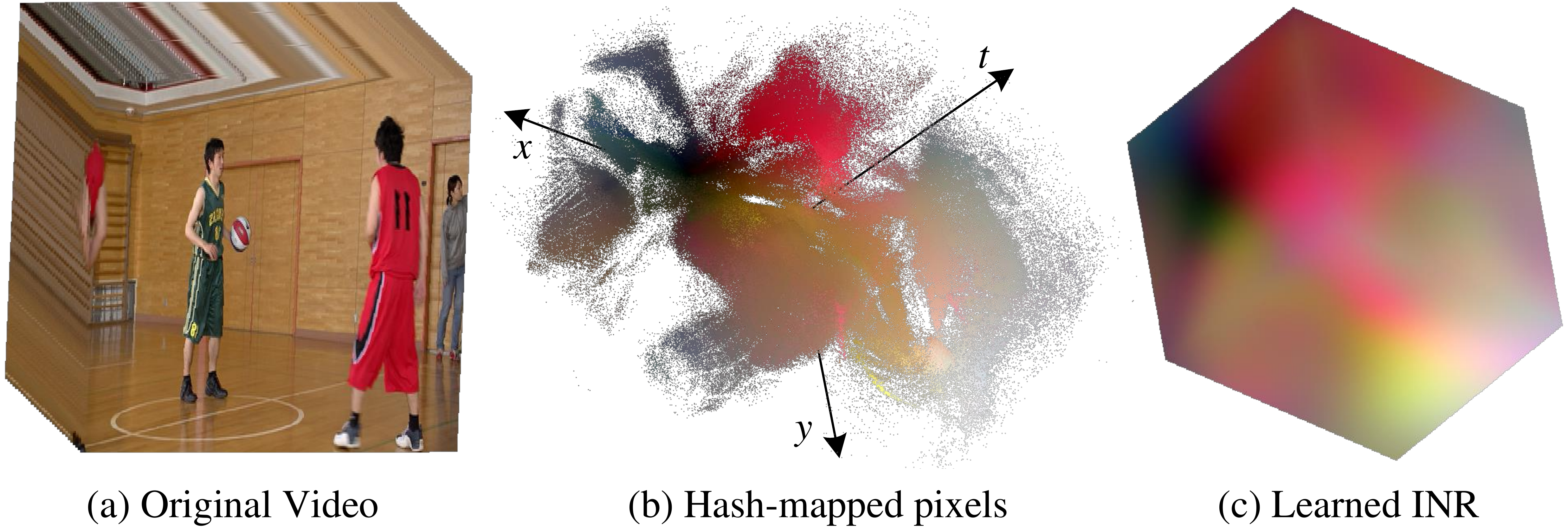}
  \caption{Visualization of learned INR on the 3D video `BasketballPass'~\cite{sullivan2012overview}. (a) and (b) compare the coordinates with and without hash-table. (c) shows the learned INR of the SIREN after the mapping by hash-table.}
  \label{fig:coords_mapped_video}
  % \vspace{-0.3cm}
\end{figure}

\begin{figure*}[t]
  \centering
  \includegraphics[width=0.9\linewidth]{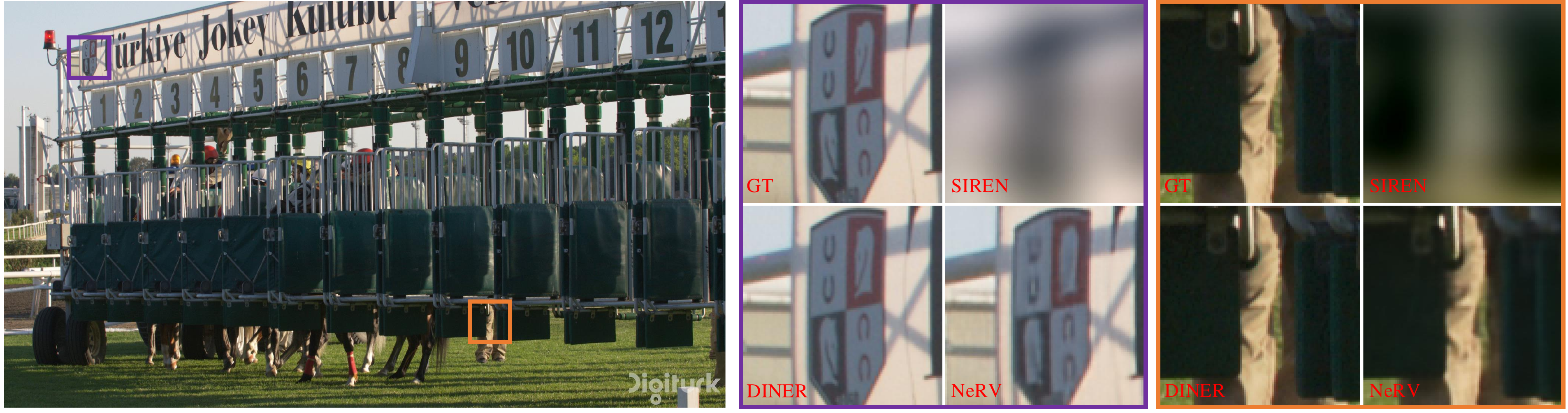}
  \caption{Qualitative comparisons of various methods on 3D video representation after 500 epochs. The proposed DINER outperforms compared baseline methods. Specifically, SIREN obtains the smoothed/blurred results, considering the tiny network compared to the high resolution. NeRV recovers the HD video but losses the high-frequency details, such as the billboard and horse leg of the right zoom-in images in details.} 
  \label{fig:video_res_cmp}
 % \vspace{-0.3cm}
\end{figure*}

\begin{table}[t]
\small
\caption{Comparisons of training 3D videos 'ReadySetGo' and 'ShakeNDry' with 500 epochs.}
  \label{tab:3D_video_psnr}
  \centering
  \begin{tabular}{@{}lccc@{}}
    \toprule
    Methods & Network Para. & Training time & PSNR  \\
    \midrule
    SIREN & 8.77K & 706s & 21.22 dB \\
    NeRV & 97.24M & 7445s & 36.08 dB\\
    DINER & 8.77K & 1309s & 50.74 dB \\
    \bottomrule
  \end{tabular}
  
     % % \vspace{-0.3cm}
\end{table}

\subsection{Comparisons with the SOTAs on Gigapixel image representation}
Mapping the coordinates to colors of a gigapixel image could tests the performance of models in high-frequency details and is an important task in INR.
We compare the DINER with two recent SOTAs, \textit{i.e.}, the ACORN~\cite{martel2021acorn} and the InstantNGP~\cite{muller2022instant} (same parameter settings as used in the above experiment) using the Pluto image. Fig.~\ref{fig:gigapixel_res_cmp} provides comparisons on the Pluto with the resolution $8000\times 8000$. DINER could provide much reliable results for gigapixel image than SOTAs. 

\subsection{Comparisons with the SOTAs on representing 3D video}
Video describes a dynamic 3D scene $I(t,x,y)$ composed of multiple frames. Accurate representation for 3D video is becoming a popular task~\cite{chen2021nerv,rho2022neural,kim2022scalable,li2022nerv} in the community of INR. We compare the proposed DINER with the SIREN~\cite{sitzmann2020implicit} and the state-of-the-art INR NeRV~\cite{chen2021nerv}. Noting only the Hash+SIREN architecture is evaluated in this task. The NeRV is implemented using their default parameters, while the SIREN and the proposed Hash+SIREN both use the same network structure with the size $4\times 64$. All three methods are evaluated on the videos 'ReadySetGo' and 'ShakeNDry' of the UVG dataset~\cite{mercat2020uvg} and are trained with 500 epochs. The first 30 frames with 1920$\times$1080 resolution are used in our experiment. 

Fig.~\ref{fig:coords_mapped_video}(a) and (b) show the mapping of the coordinates with and without mapping. Fig.~\ref{fig:coords_mapped_video}(c) illustrates the learned INR. It is noticed that the low-frequency property also appears in the learned INR of the 3D video in our DINER. Tab.~\ref{tab:3D_video_psnr} shows the quantitative comparisons. The proposed method outperforms the NeRV both in quality and speed with $14$ dB and $5\times$ improvements, respectively. Because a large hash-table is used in DINER, more time are taken in the transmission between the memory and the cache in the GPU, resulting more training time of DINER than the SIREN. Fig.~\ref{fig:video_res_cmp} shows the qualitative comparisons. Noting that the SIREN with a tiny network could not provide reasonable representation for the `ReadySetGo' data with $30\times 1920\times 1080$ pixels, thus all pixels are smoothed. NeRV provides better results than the ones by SIREN, however the high-frequency details are lost such as the character `U' (left-top corner) and the red logo of horsehead (right-top corner) in the purple box, as well as the folds of the trousers in the orange box. On the contrary, the original video is mapped with little high-frequency component in the proposed DINER (Fig.~\ref{fig:coords_mapped_video}(c)). As a result, the details mentioned above could be well represented.

\section{Applications}
In this section, we will verify the performance of the DINER on solving inverse problems, three separate tasks are conducted, \textit{i.e.}, 2D phase retrieval in lensless imaging, 3D Refractive Index recovery in intensity diffraction tomography, and neural radiance field optimization. Note that, the width of the hash-table for these tasks are set according to the analysis in Sec.~\ref{sec:diner_expressivepower}, thus some results may be different from the conference version~\cite{xie2023diner}.

\subsection{Phase Recovery in Lensless Imaging}
Lensless imaging~\cite{ozcan2016lensless} observes specimen in a very close distance without any optical lens. By directly recording the diffractive measurements, it provides the advantage of wide field of view observation and has become an attractive microscopic technique~\cite{zhou2020wirtinger} for analyzing the properties of the specimen. We take the classic multi-height lensless imaging as an example, where $N$ measurements $\{I_z\}_{z=z_1}^{z_N}$ are captured under different specimen-to-sensor distances $z$ for the specimen's amplitude and phase imaging recovery. In multi-height lensless imaging, $I_{z}$ could be modeled as applying Fresnel propagation to the complex field $O(x,y)$ of the specimen, \textit{i.e.},
\begin{equation}
\label{eqn:inverse_lensless}
    I_z = |PSF_{z}*[P(x,y)\cdot O(x,y)]|^2,
\end{equation}
where $P(x,y)$ is the illumination pattern, $PSF_{z}$ is the point spread function of Fresnel propagation over distance $z$ between the specimen and the sensor.

\begin{figure}[t]
  \centering
  \includegraphics[width=0.68\linewidth]{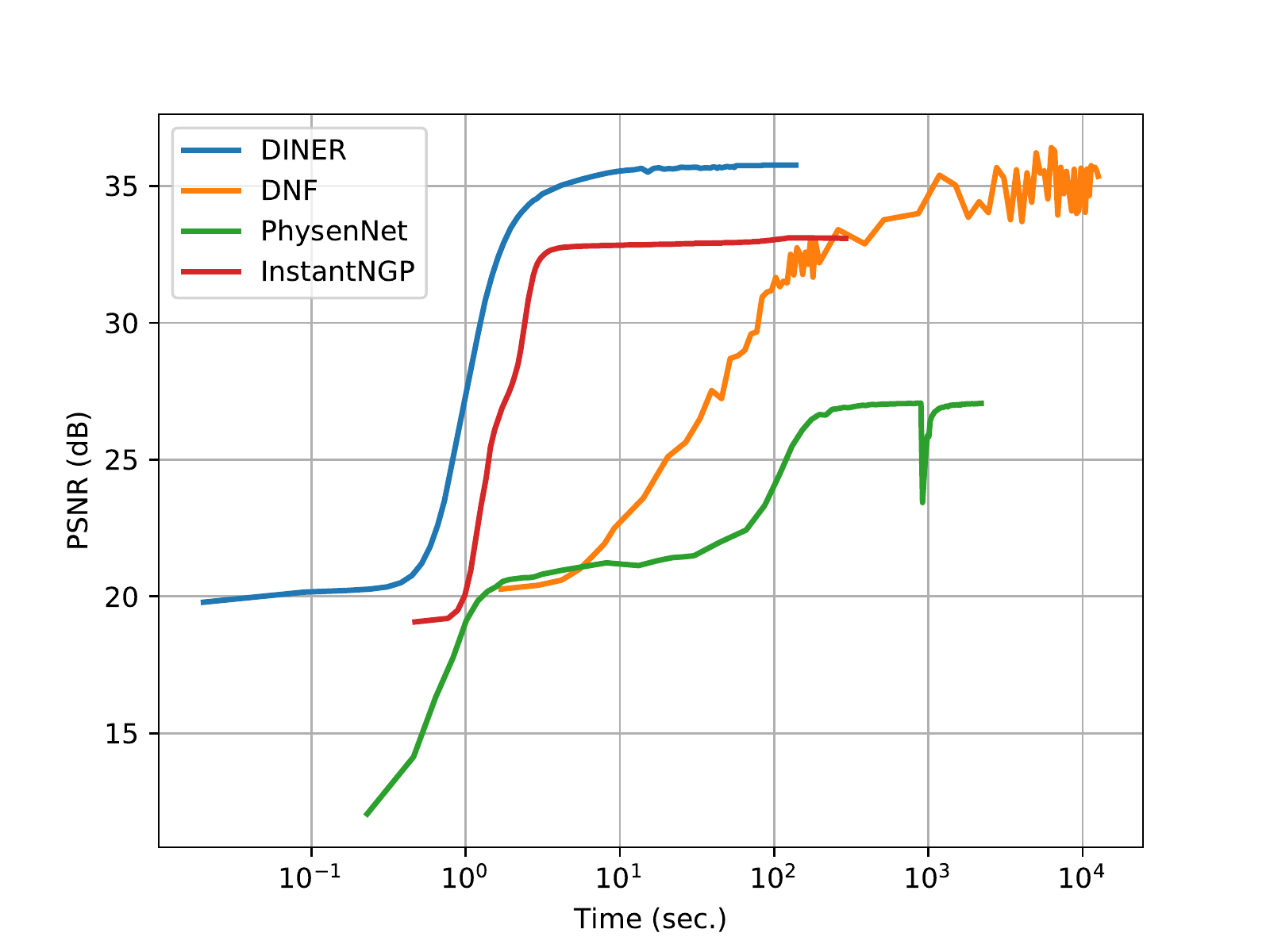}
  \caption{PSNR of reconstructed measurements over training time on the real data of lensless imaging.}
  \label{fig:lensless_PSNR_over_epochs:real}
  % \vspace{-0.3cm}
\end{figure}

\begin{figure}[t]
  \centering
  \includegraphics[width=\linewidth]{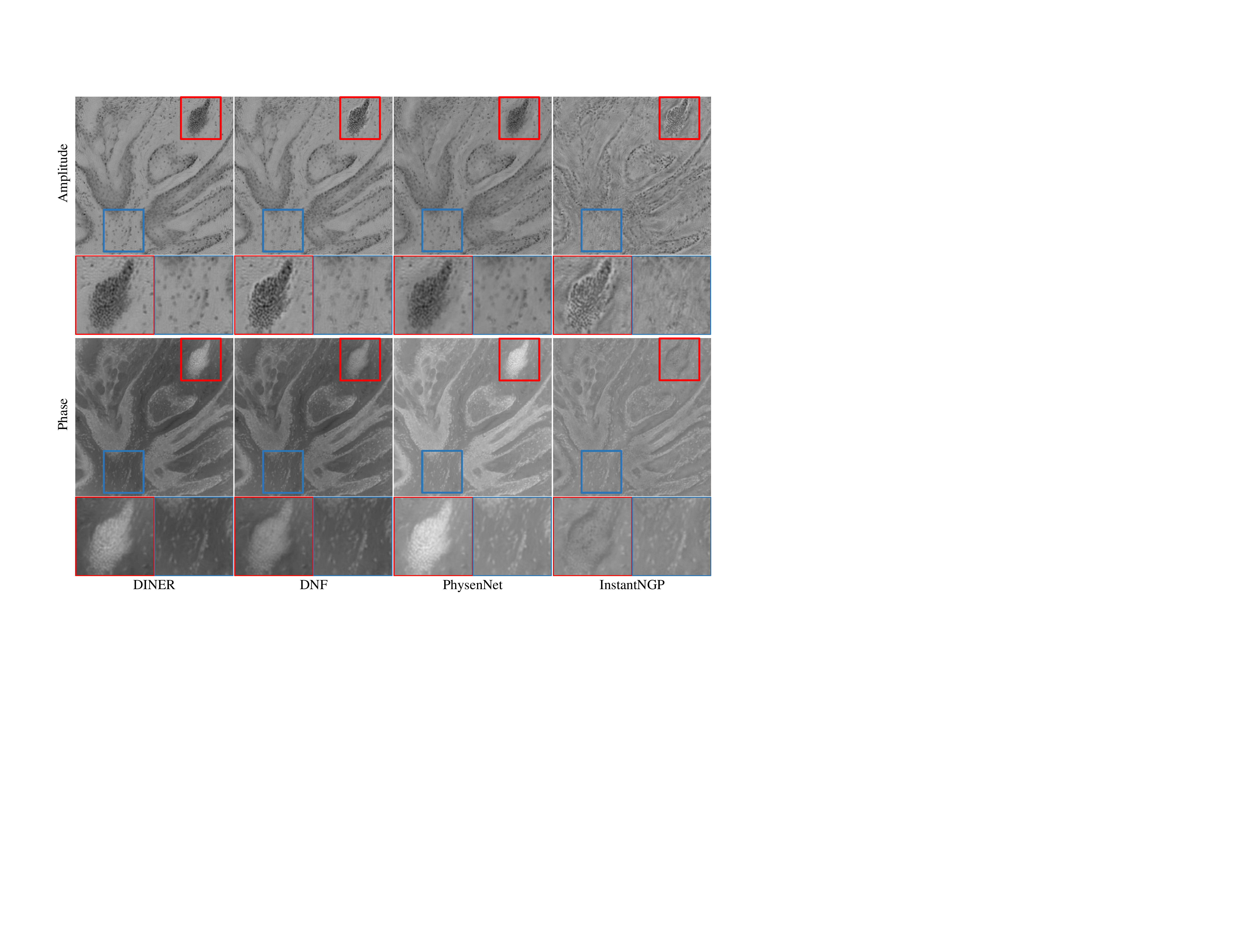}
  \caption{Comparisons on real lensless imaging of the animal skin section. While having reconstruction performance close to DINER, DNF and PhysenNet take $100\times$ as much computational time to converge. InstantNGP achieves fast training speed, but its results contain heavy artifacts. In contrast, the proposed DINER takes less time to converge and recovers high-quality details of the animal skin section.}
  \label{fig:cmp_lensless:real}
\end{figure}

We model $O(x,y)$ using the proposed method with the SIREN backbone and the network size is $2\times 64$. The loss function is built by comparing the measurements with the results from applying the Eqn.~\ref{eqn:inverse_lensless} to the network output. We compare our method with the current SOTAs, \textit{i.e.}, the diffractive neural field (DNF)~\cite{zhu2022dnf} (the backbone is the PE+MLP structure) and the PhysenNet~\cite{wang2020phase}, additionally, the InstantNGP~\cite{muller2022instant} is also applied in the lensless imaging model (with similar hash-table parameters setting in the DINER) and compared. Note that only the results on real measurements are provided here, please refer the conference version for comparisons on the synthetic data.

% Fig.~\ref{fig:lensless_PSNR_over_epochs:syn} shows the PSNR curves of reconstructed measurements over training time on the synthetic data. The proposed method has $18\times$ and $80\times$ advantages on the convergence speed over the PhysenNet and DNF, respectively. The InstantNGP has similar convergence speed with the DINER, however the later has a $18$dB advantage on the PSNR metric. Fig.~\ref{fig:cmp_lensless} provides qualitative comparisons of the reconstructed amplitude and phase. Although only $1.5\%$ network parameters ($2\times 64$ \textit{vs.} $8\times 256$) are used, the proposed DINER could provide better results than the DNF thanks to the hash-table. 

Fig.~\ref{fig:lensless_PSNR_over_epochs:real} shows the PSNR curves of reconstructed measurements on the real data of animal skin section. DINER achieves higher PSNR values ($>7$ dB) on the reconstructed measurements than the PhysenNet. Additionally, DINER has about $100\times$ improvement in the convergence speed when arriving $25$ dB. Although the DNF could achieve similar results with DINER, it takes more than $100\times$ training times to arrive $35$ dB since a complex MLP is used ($8\times 256$ \textit{vs.} $2\times 64$). The InstantNGP has similar convergence speed with the DINER, however the later has a $3$dB advantage on the PSNR metric. Fig.~\ref{fig:cmp_lensless:real} shows qualitative results of different methods  conducting the same epochs ($10000$ times). DINER can resolve fine details of the skin sample. Although DNF and PhysenNet achieve similar reconstruction performance with DINER, but they use about $100\times$ computational time. The imaging quality by InstantNGP is relatively low with heavy reconstruction artifacts, although its convergence speed is as fast as DNIER. InstantNGP cannot accurately recover the high-resolution image details, as the zoom-in images shown in Fig.~\ref{fig:cmp_lensless:real}.

\subsection{3D Refractive Index Recovery in Intensity Diffraction Tomography}
The 3D refractive index characterizes the interaction between light and matter within a specimen. It is an endogenous source of optical contrast for imaging specimen without staining or labelling, and plays an important role in many areas, \textit{e.g.} the morphogenesis, cellular pathophysiology, biochemistry~\cite{park2018quantitative}. Intensity diffraction tomography measures the squared amplitude of the light scatted from the specimen at different angles multiple times and has become a popular technique for recovering the 3D refractive index. 

\begin{figure}[t]
  \centering
  \includegraphics[width=\linewidth]{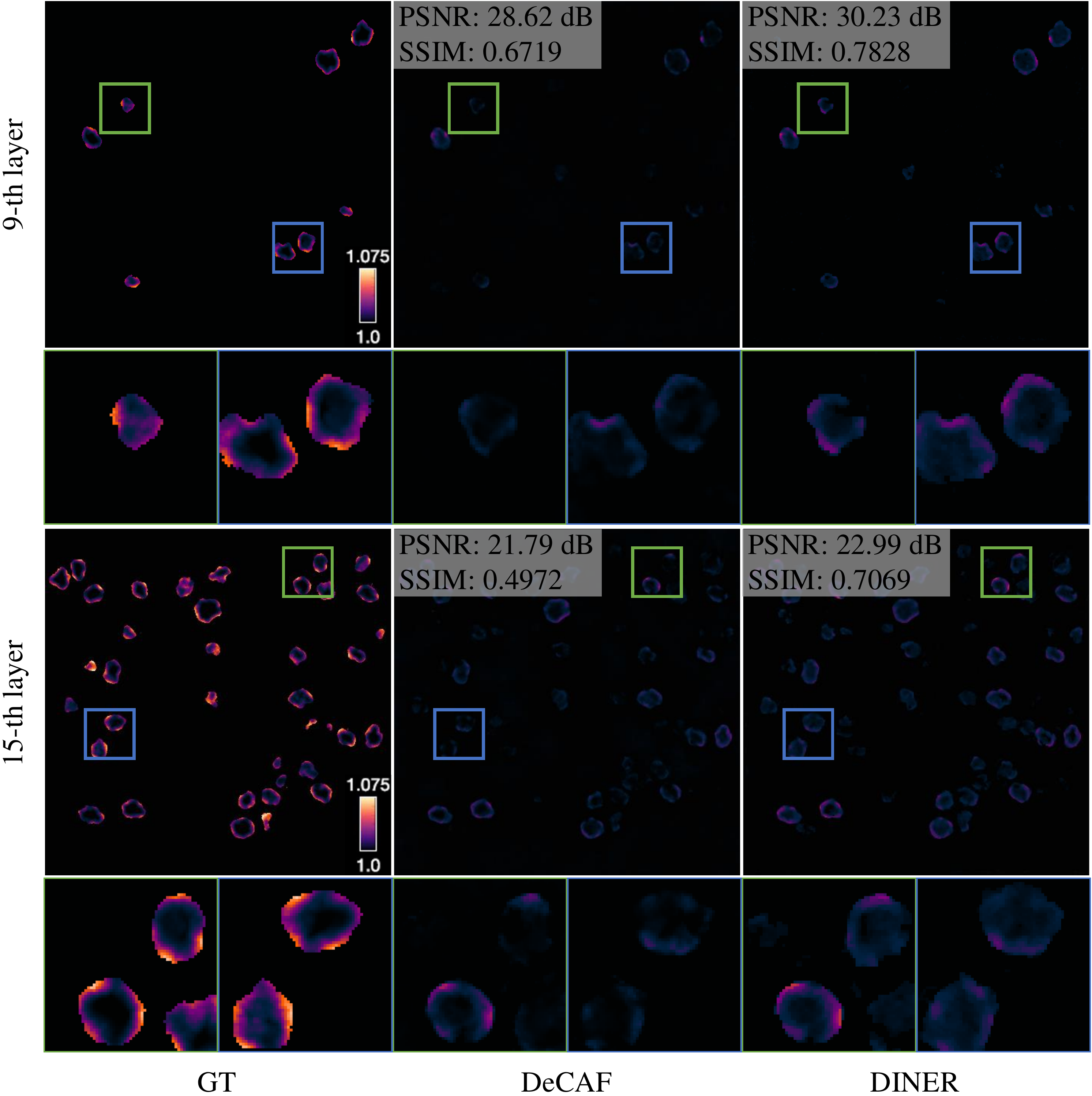}
  % \vspace{-0.2cm}
  \caption{Comparisons on 3D refractive index recovery. DINER takes less training time and could reconstruct more surface details of the Granulocyte.}% \vspace{-0.5cm}
  \label{fig:cmp_RI}
\end{figure}

Given the 3D refractive index $\mathbf{n} = (\mathbf{n}_{re} + j\mathbf{n}_{im})$ of a specimen, where $\mathbf{n}_{re}$ and $\mathbf{n}_{im}$ are the real and imaginary parts of the specimen's refractive index, respectively. The forward imaging process of sensor placed at location $\rho$ could be modeled as 
\begin{equation}
\label{eqn:IDT_forward}
I_{\rho} = \mathbf{A}_{\rho}\Delta \epsilon,
\end{equation} 
where $\mathbf{A}_{\rho}$ records the sample-intensity mapping with the illuminations. $\Delta \epsilon=\Delta \epsilon_{re} + j\Delta \epsilon_{im}$ is the complex-valued permittivity contrast and could be obtained by solving
\begin{equation}
\begin{aligned}
&\mathbf{n}_{\mathrm{re}}=\sqrt{\frac{1}{2}\left(\left(\mathbf{n}_0^2+\Delta \epsilon_{\mathrm{re}}\right)+\sqrt{\left(\mathbf{n}_0^2+\Delta \epsilon_{\mathrm{re}}\right)^2+\Delta \epsilon_{\mathrm{im}}^2}\right)} \\
&\mathbf{n}_{\mathrm{im}}=\frac{\Delta \epsilon_{\mathrm{im}}}{2 \cdot \mathbf{n}_{\mathrm{re}}}
\end{aligned},
\end{equation}
where $\mathbf{n}_0$ is the refractive index of the background medium.

We model the $(\Delta \epsilon_{re}, \Delta \epsilon_{im})$ using the DINER with network size $2\times 64$ and the same loss function is used as the DeCAF. We compare our method with the SOTA, \textit{i.e.}, DeCAF~\cite{liu2022recovery} which uses a combination of the standard MLP structure with network size $10\times 208$ and positional+radial encodings.

\begin{figure*}[t]
  \centering
  \includegraphics[width=0.9\linewidth]{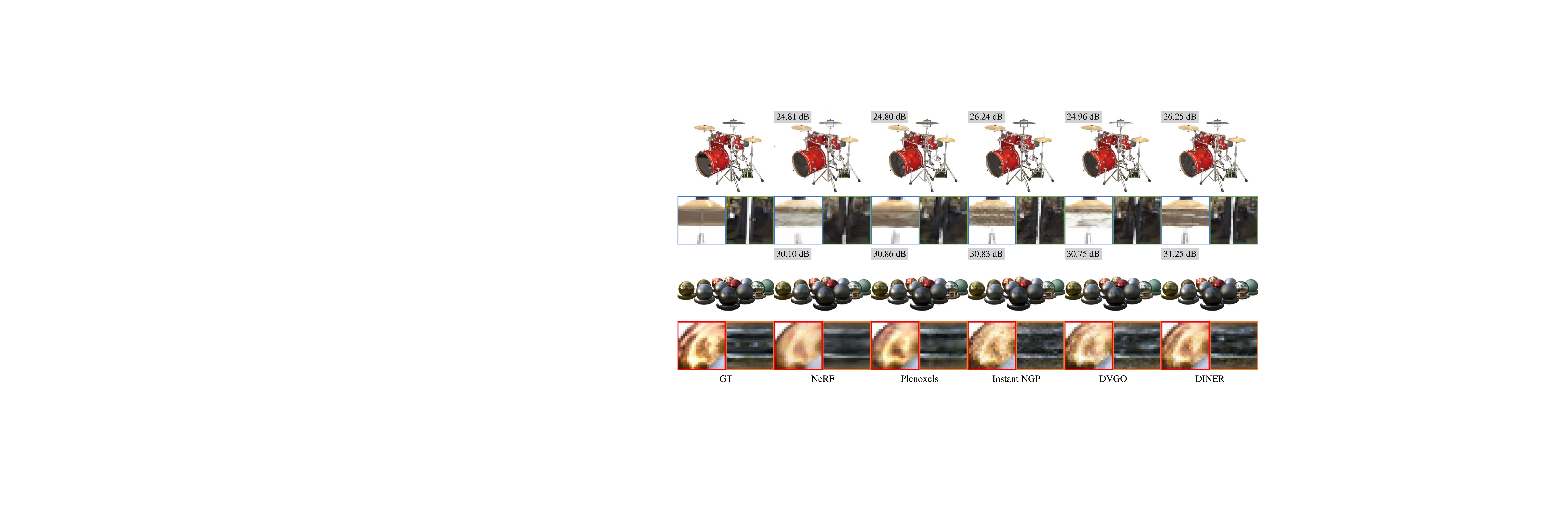}
  \caption{Qualitative comparisons of various methods on the task of novel view synthesis.}
  \label{fig:nerf_res_all}
  % \vspace{-0.2cm}
\end{figure*}

Fig.~\ref{fig:cmp_RI} compares our method with the DeCAF on the 3D Granulocyte Phantom data using the Fiji software~\cite{schindelin2012fiji}. Because the ground truth of the Granulocyte Phantom is released recently, we modify the weights of different components in the loss function, results in different images in Fig.~\ref{fig:cmp_RI} with the conference version~\cite{xie2023diner}. To achieve the above results, the DeCAF takes $401$ minutes while the DINER only takes $91$ minutes. Since the hash-table could map a high-frequency signal in a low-frequency way, our results provide more details on the surface of the Granulocyte. While the surface boundaries of the Granulocyte are over-smoothed by the background in the results of the DeCAF since the PE+MLP could not accurately model the high-frequency components. %(Please refer to the supplemented video on different heights for better comparison).

\subsection{Neural Radiance Field Optimization}
\label{sec:exp_nerf}

\definecolor{red}{rgb}{1.0000,0.5686,0.5059}
\definecolor{orange}{rgb}{1.0000,0.7373,0.5059}
\definecolor{yellow}{rgb}{1.0000,0.8431,0.5059}

\begin{table}
  \centering
  \caption{Quantitative comparisons on novel view synthesis. The values in `average' error metric refer to the average of the mean square error, $\sqrt{1-SSIM}$ and LPIPS~\cite{barron2021mip}.}
  \label{tab:psnr_nerf}
  \begin{tabular}{@{}lccccc@{}}
    \toprule
     & DINER & NeRF & Plenoxels & InstantNGP & DVGO \\
    \midrule
    PSNR & \cellcolor{yellow}33.18  & 31.04 & 30.99 & \cellcolor{red}33.53 & \cellcolor{orange}33.38\\
    SSIM & \cellcolor{orange}0.967 & 0.953 & 0.956 & \cellcolor{yellow}0.964 & \cellcolor{red}0.968\\ 
    LPIPS & \cellcolor{orange}0.033 & 0.054 & 0.050 & \cellcolor{yellow}0.038 & \cellcolor{red}0.032\\
    Average & \cellcolor{orange}0.067 & 0.088 & 0.084 & \cellcolor{yellow}0.072 & \cellcolor{red}0.067\\
    Time (sec.) & 418 & 10887 & \cellcolor{orange}305 & \cellcolor{yellow}343 & \cellcolor{red}252 \\
    \bottomrule
  \end{tabular}
\end{table}

Neural radiance field (NeRF) promotes the development of the novel view synthesis a lot. Given a sparse set of images captured from different positions, the task of novel view synthesis aims at synthesizing images from the positions that are not exist in the input set. NeRF achieves the SOTA performance compared with traditional depth-related methods (\textit{e.g.}, the explicit-depth-based methods, the no-depth-based methods and the half-depth-based methods). One of the most successful experiences in NeRF is the usage of the INR, which models the attributes of a light ray in 3D space as a neural network. NeRF takes the position $\vec{x}$ and direction $\vec{d}$ of each ray as the INR input and outputs the color $c$ and transmission $\sigma$. After that, the ray marching process is applied to the INR output to achieve the rendering of the color for each pixel. Finally, a loss function between the rendered color $\hat{C}$ and the ground truth $C$ is built to supervise the training of the INR. The whole progress could be written in the following formulas
\begin{equation}
\begin{aligned}
& (c_i,\sigma_i)=INR(\vec{x}_i,\vec{d}_i)\\
& \hat{C}(p) = \sum_{i}T_i(1-e^{-\sigma_i\delta_i})c_i,\:\: T_i=e^{-\sum_{j<i}\sigma_j\delta_j}\\
& loss = \sum_{p}||C(p)-\hat{C}(p)||_{2}^{2},
\end{aligned}
\end{equation}
where $\delta_i$ refers to the interval between the neighbouring sampling points along a light ray and is often a constant value.

\begin{figure*}[h]
\begin{center}
\centering
\subfloat[PSNR with different hash-table widths]{
	\label{fig:redundancy_SH_coefficients:hash_table_width}
	\includegraphics[width=0.35\linewidth]{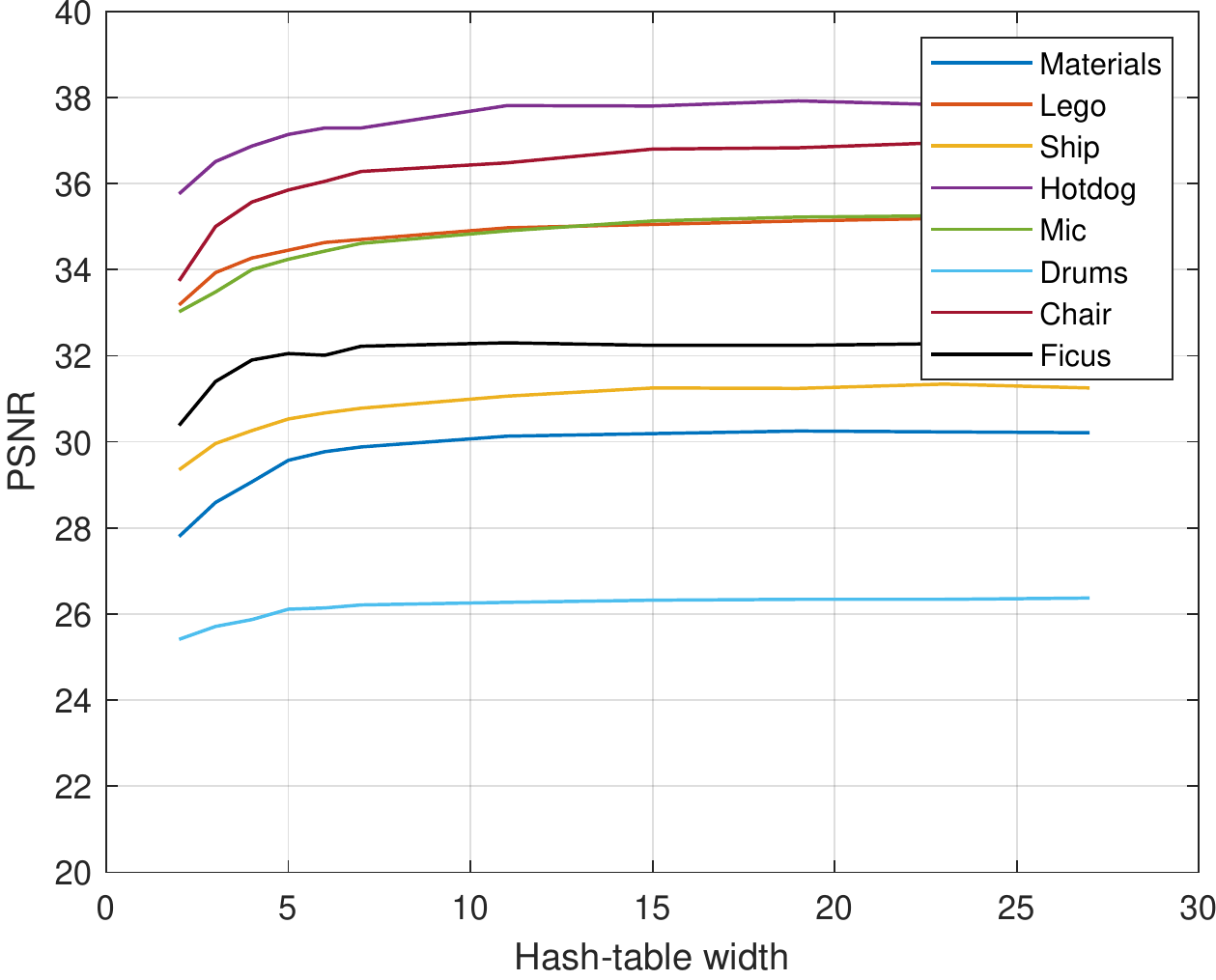}
}
\subfloat[Contributions of the top-$K$ principle components in SH coefficients]{
	\label{fig:redundancy_SH_coefficients:pca_analysis}
	\includegraphics[width=0.355\linewidth]{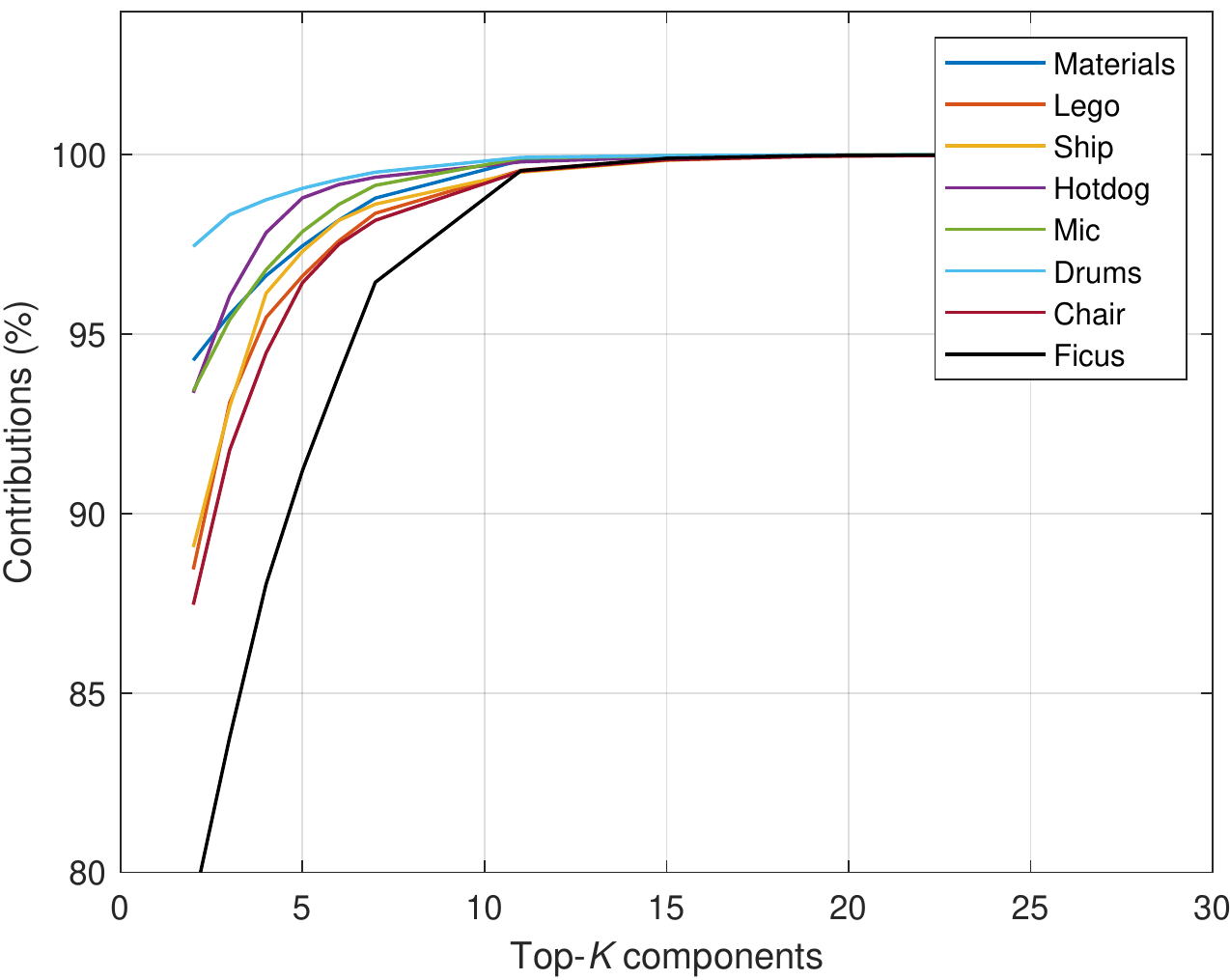}
}
\end{center}
\caption{The redundancy of the SH coefficients. (a) The PSNR of novel view synthesis with different hash-table widths. (b) The contributions of the top-$K$ principle components by analyzing the SH coefficients using the PCA~\cite{bishop2006pattern}.}
\label{fig:redundancy_SH_coefficients}
\end{figure*}
Due to the complex structure and texture distribution of the scene, a large MLP ($8\times 256$ network structure and $10$ Fourier bases in the PE) is used in NeRF, resulting in a long training time. DINER could significantly accelerate the training of the NeRF using a small MLP network with the size $1\times 128$. Because the DINER could only process the discrete signals (please refer the conference version or the Fig.~\ref{fig:limitation} for the discussion of the continuous signals), we split the continuous 3D world as a voxel grid with $160^3$ points and adopt the spherical harmonic (SH) coefficients~\cite{yu2021plenoctrees,fridovich2022plenoxels} to model the color variance of the grid points observed from different directions. The DINER-based novel view synthesis is built upon the code of the direct voxel go optimization (DVGO)~\cite{sun2022direct}. We compare the DINER based novel view synthesis with the original NeRF and three recent SOTAs, \textit{i.e.}, the InstantNGP~\cite{muller2022instant}, the Plenoxels~\cite{fridovich2022plenoxels} and the DVGO~\cite{sun2022direct}. 

Tab.~\ref{tab:psnr_nerf} and Fig.~\ref{fig:nerf_res_all} provide the quantitative and qualitative results on the down-scaled Blender dataset~\cite{mildenhall2020nerf} with $400\times 400$ resolution, respectively. Compared with Plenoxels which adopts the same representation (voxel grid and SH coefficients), DINER provides better performance for optimizing SH coefficients than the non-neural network based method used in the Plenoxels. Although the expressive ability of the SH coefficients is weaker than the continuous way~\cite{Karnewar2022relufield}, DINER-based method could provide better results than the original NeRF and competitive results with the InstantNGP and the DVGO, verifying good expressive power of the DINER.

\noindent \textbf{Redundancy of the SH coefficients.} Yu et al.~\cite{yu2021plenoctrees} suggested setting the SH coefficients as a $27$-dimensional vector to model the view-dependent appearance. We noticed that it is redundant to use such a $27$-dimensional coefficients by analyzing the performance of the novel view synthesis with different hash-table widths. 

% We experimental verify the performance of the DINER with different hash-table widths. Two data, \textit{i.e.}, the Hotdog and the Materials, which contain the most and the least view-dependent appearance are selected for verification. 

Fig.~\ref{fig:redundancy_SH_coefficients}(a) plots the PSNR with different hash-table widths. It is noticed that the values increase a lot at beginning and tends to be stable when the width reaches about $15$. According to the analysis in Sec.~\ref{sec:exp_powers_width} and the results in Sec.~\ref{sec:res_2dimg_dif_width}, the trend of curves in Fig.~\ref{fig:redundancy_SH_coefficients}(a) indicates that the SH coefficients is rank-deficient. To verify this conclusion, we apply the principle component analysis (PCA)~\cite{bishop2006pattern} to the output $27$-dimensional SH coefficients when the hash-table width is set as $27$. PCA is a dimensionality-reduction tool by selecting the top-$K$ linear-independent principle components in the transformed domain of the signal, as a result, the contributions of the selected top-$K$ components could reflect the linear-dependency of the signal~\cite{bishop2006pattern}. Fig.~\ref{fig:redundancy_SH_coefficients}(b) plots the contributions of the first top-$K$ dimensions in PCA. It is noticed that the curves of contributions in Fig.~\ref{fig:redundancy_SH_coefficients}(b) have similar tendency with the PSNR curves in Fig.~\ref{fig:redundancy_SH_coefficients}(a) (\textit{i.e.}, both the PSNR and the contribution values increase at the beginning and tend to be stable when the hash-table width or the top-$K$ components reach about $15$),  verifying the redundancy of SH coefficients concluded by analyzing the performance with different hash-table width (\textit{i.e.}, the Fig.~\ref{fig:redundancy_SH_coefficients}(a)).

% \begin{table*}
%   \centering
%   \caption{Quantitative comparisons on novel view synthesis. Values here refer to the `average' error metric presented in \cite{barron2021mip} (\textit{i.e.}, the average of the mean square error, $\sqrt{1-SSIM}$ and LPIPS).}
%   \label{tab:psnr_nerf}
%   \begin{tabular}{@{}lccccccccc@{}}
%     \toprule
%      & Materials & Hogdog & Ship & Mic & Drums & Lego & Chair & Ficus & Mean\\
%     \midrule
%     DINER \\
%     NeRF \\
%     Plenoxels \\
%     InstangNGP \\
%     DVGO \\
%     \bottomrule
%   \end{tabular}
% \end{table*}

\begin{figure}[t]
  \centering
  \includegraphics[width=\linewidth]{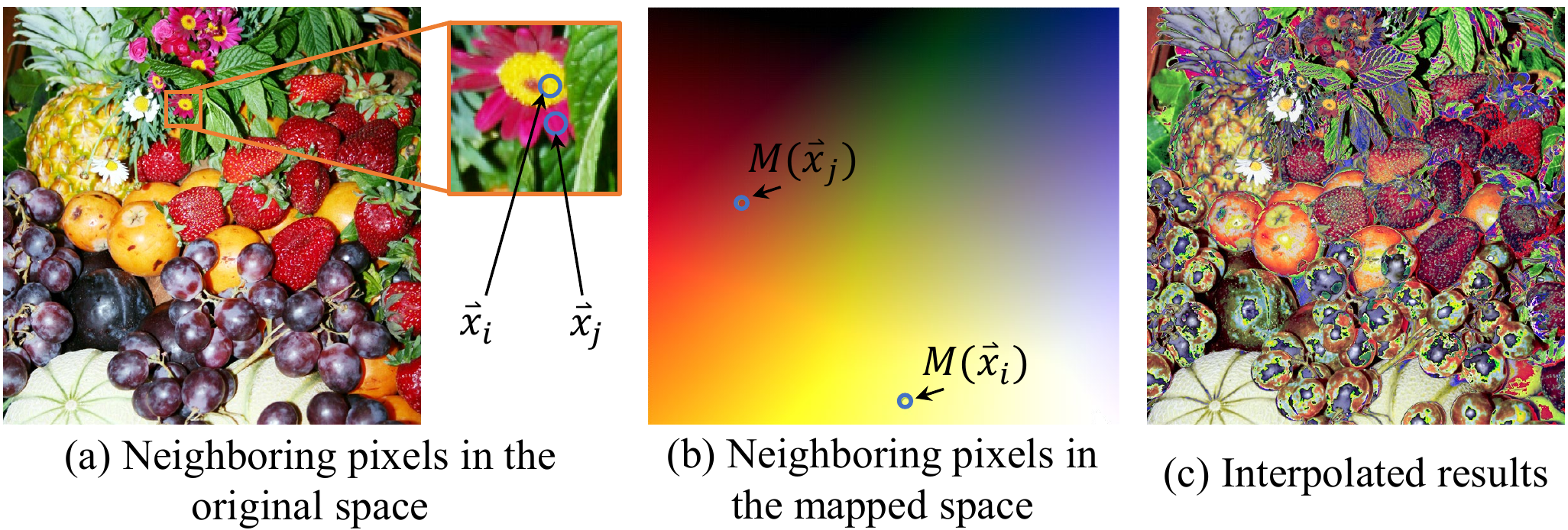}
  % \vspace{-0.4cm}
  \caption{Analysis of feeding interpolated hash-key to the DINER. (a) Two neighboring coordinates $\vec{x}_i$ and $\vec{x}_j$ are labelled in the original image. (b) The distance between the mapped coordinates $\mathcal{HM}(\vec{x}_i)$ and $\mathcal{HM}(\vec{x}_j)$ is larger than the one in the original space. (c) Results by feeding the interpolated mapped coordinates to the trained MLP.}% \vspace{-0.5cm}
  \label{fig:limitation}
\end{figure}
\subsection{Discussion}
The DINER is designed for discrete signals. To query an unseen coordinate (\textit{i.e.}, no corresponding hash-key in the hash-table) in a continuous signal,  
it is suggested to apply a post-interpolation operation to the network output instead of feeding interpolated hash-key to the network (see Fig. \ref{fig:limitation}), such as the work in Sec.~\ref{sec:exp_nerf} which splits the continuous radiance field as a voxel grid and models the color variance using spherical harmonic coefficients~\cite{fridovich2022plenoxels} instead of feeding unseen position and direction coordinates to the network directly~\cite{mildenhall2020nerf}.

\section{Conclusion}
In this work, we have proposed the DINER which could greatly improve the accuracy of current INR backbones by introducing an additional hash-table. We have pointed out that the performance of INR for representing a signal is determined by the arrangement order of elements in it. The proposed DINER could map the input discrete signal into a low-frequency one, which is invariant if only the arrangement order changes while the histogram of attributes is not changed. For this reason, the accuracy of different INR backbones could be greatly improved. Additionally, we have proved that the expressive power of the DINER could be greatly improved by increasing the width of the hash-table until the rank of the signal. Extensive experiments have verified the high accuracy and efficiency of the proposed DINER for tasks of signal fitting and inverse problem optimization.

However, the current DINER could only process discrete signals. In the future, we will focus on continuous mapping methods instead of discrete hash-table-based mapping to extend the advantages for continuous signals such as the signed distance function~\cite{park2019deepsdf}.

% or


\appendices
\section{Proof of the Eqn.~\ref{eqn:hyper_surface_set:1}}
Supposing the parametric variable $x=[x_1,x_2,x_3]^{\top}$ is a 3D coordinate, the `Hyper-curved-surface' in $\mathcal{S}_3^{\omega}$ could be described as
\begin{equation}
\small
\left\{
\begin{aligned}
r&=\sum_{\omega'\in\mathcal{H}_{\Omega}}c_{\omega'}^{1}\sin (\langle\omega'
,[x_1,x_2,x_3]^{\top}\rangle+\phi_{\omega'}^{1})\\
g&=\sum_{\omega'\in\mathcal{H}_{\Omega}}c_{\omega'}^{2}\sin (\langle\omega'
,[x_1,x_2,x_3]^{\top}\rangle+\phi_{\omega'}^{2})\\
b&=\sum_{\omega'\in\mathcal{H}_{\Omega}}c_{\omega'}^{3}\sin (\langle\omega'
,[x_1,x_2,x_3]^{\top}\rangle+\phi_{\omega'}^{3})
\end{aligned}
\right.,
\end{equation}
where $\omega'=[\omega_1',\omega_2',\omega_3']^{\top}$ is also a 3D vector here. By setting the $x_3=0$, the above functions is simplified to the following one
\begin{equation}
\small
\left\{
\begin{aligned}
r&=\sum_{\omega'\in\mathcal{H}_{\Omega}}c_{\omega'}^{1}\sin (\langle\omega'
,[x_1,x_2,0]^{\top}\rangle+\phi_{\omega'}^{1})\\
g&=\sum_{\omega'\in\mathcal{H}_{\Omega}}c_{\omega'}^{2}\sin (\langle\omega'
,[x_1,x_2,0]^{\top}\rangle+\phi_{\omega'}^{2})\\
b&=\sum_{\omega'\in\mathcal{H}_{\Omega}}c_{\omega'}^{3}\sin (\langle\omega'
,[x_1,x_2,0]^{\top}\rangle+\phi_{\omega'}^{3})
\end{aligned}
\right.,
\end{equation}
which is equal to the parametric functions of the `Hyper-curved-surface' in $\mathcal{S}_2^{\omega}$. Consequently, 
\begin{equation}
\mathcal{S}_2^{\omega} \subset \mathcal{S}_3^{\omega},
\end{equation}
and so on, the Eqn.~\ref{eqn:hyper_surface_set:1} holds.

% \section{Proof of the Eqn.~\ref{eqn:color_space_para_funcs:conclusion_lin_depend}}
% To prove the Eqn.~\ref{eqn:color_space_para_funcs:conclusion_lin_depend}, we use the symbols as defined in Sec.~4.1. The $d_{out}$-dimensional signal attribute $\Vec{y}_i$ could be written as 

% % use section* for acknowledgment
% \ifCLASSOPTIONcompsoc
%   % The Computer Society usually uses the plural form
%   \section*{Acknowledgments}
% \else
%   % regular IEEE prefers the singular form
% \fi


\ifCLASSOPTIONcaptionsoff
  \newpage
\fi



\bibliographystyle{IEEEtran}
\bibliography{IEEEabrv,./egbib}

%\begin{thebibliography}{1}
%\end{thebibliography}

% \begin{IEEEbiography}[{\includegraphics[width=1in,height=1.25in,clip,keepaspectratio]{mshell}}]{Michael Shell}

\begin{IEEEbiography}[{\includegraphics[width=1in,height=1.25in,clip,keepaspectratio]{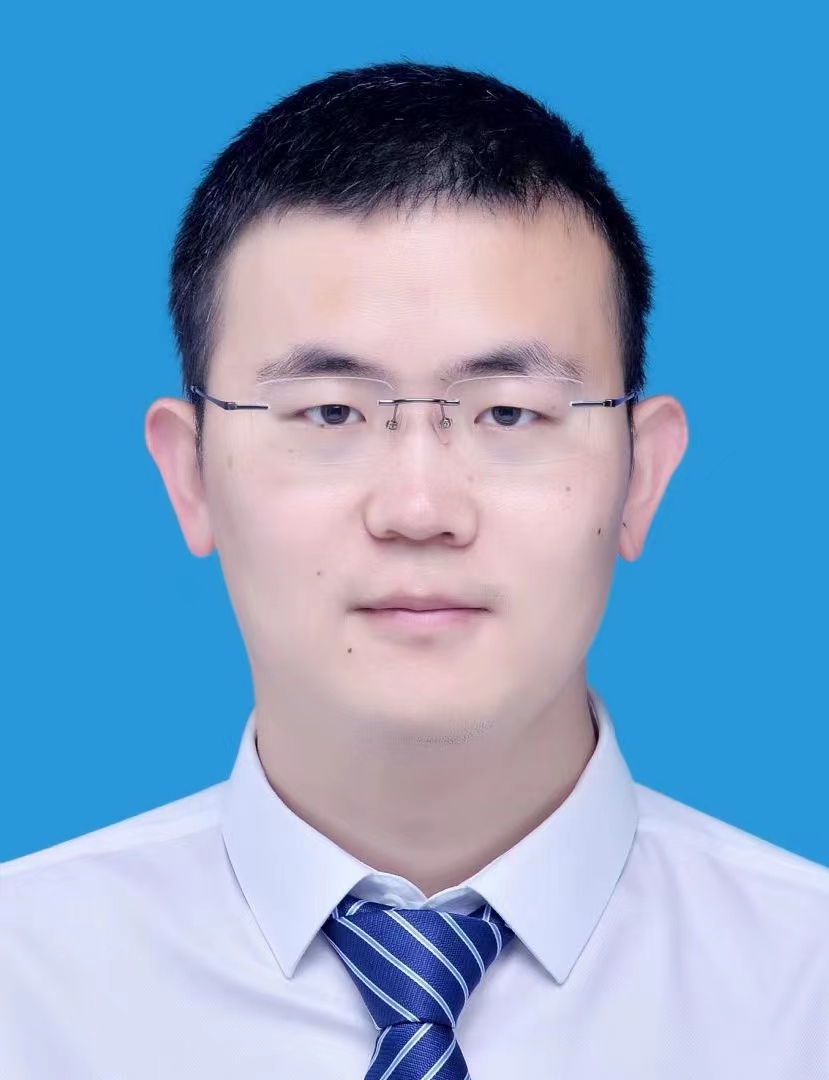}}]{Hao Zhu}
is an Associate Researcher in the School of Electronic Science and Engineering, Nanjing University. He received the B.S. and Ph.D. degrees from Northwestern Polytechnical University in 2014 and 2020, respectively. He was a visiting scholar at the Australian National University. His research interests include computational photography and optimization for inverse problems.
\end{IEEEbiography}

\begin{IEEEbiography}[{\includegraphics[width=1in,height=1.25in,clip,keepaspectratio]{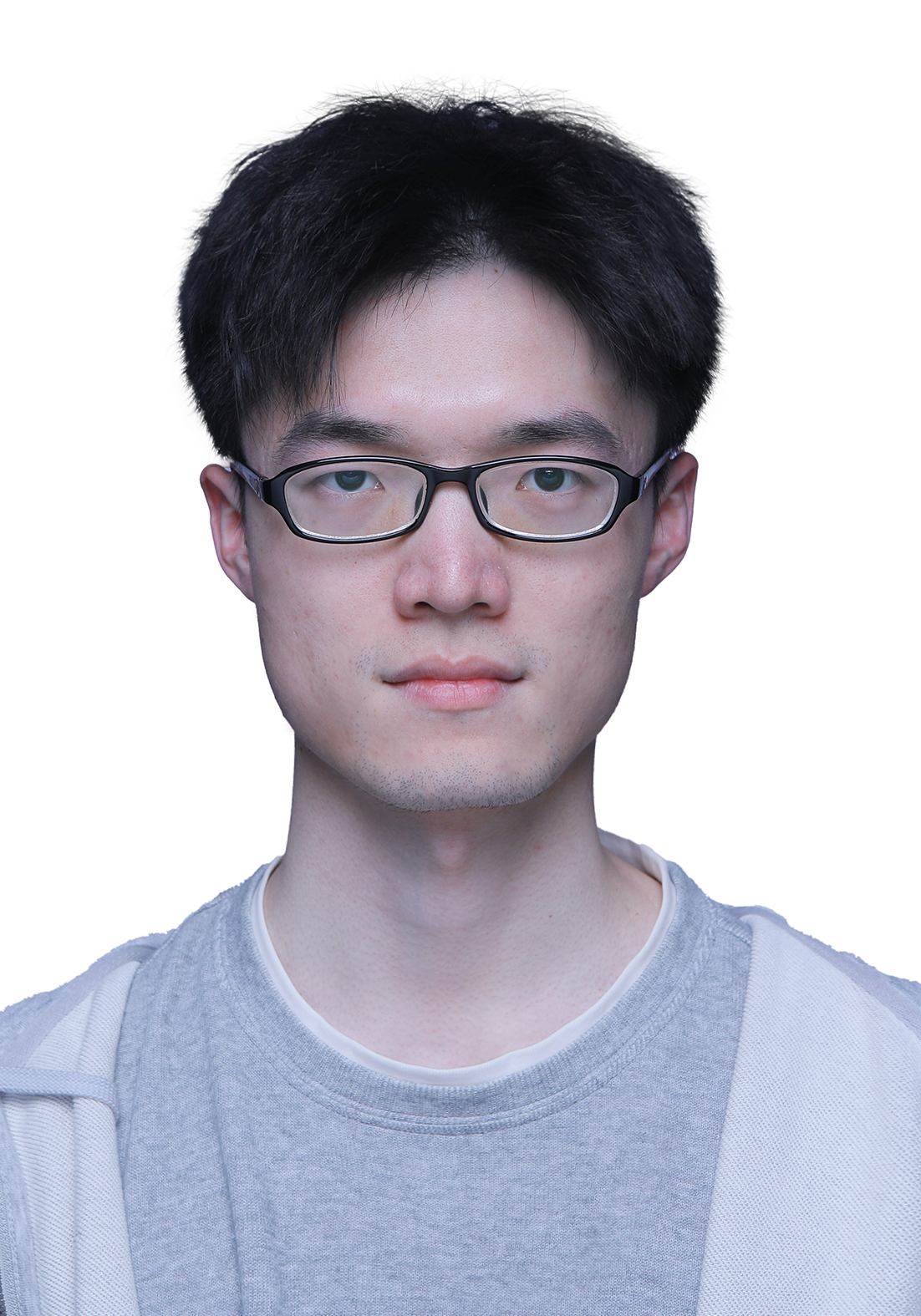}}]{Shaowen Xie} is a graduate student in the School of Electronic Science and Technology, Nanjing University. He received the B.S. degree from Nanjing University in 2021. His research interests include deep learning and implicit neural representation.
\end{IEEEbiography}

\begin{IEEEbiography}[{\includegraphics[width=1in,height=1.25in,clip,keepaspectratio]{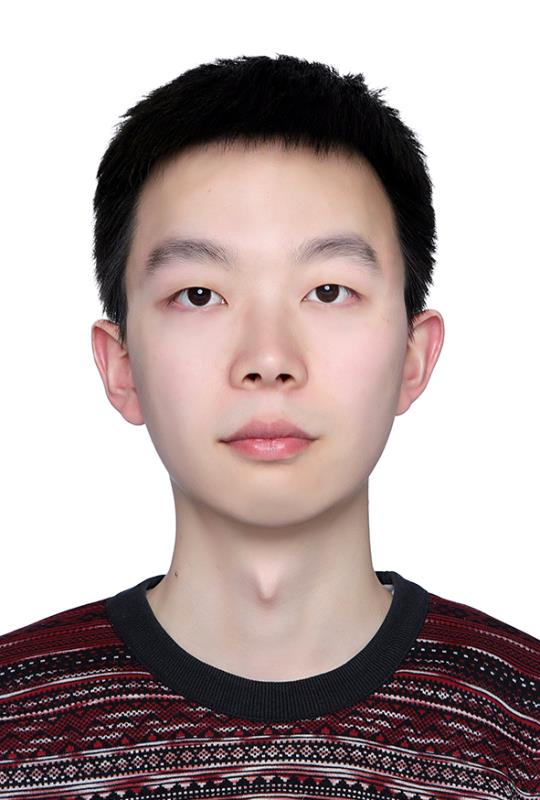}}]{Zhen Liu} is a graduate student in the Department of Computer Science and Technology, Nanjing University. He is co-supervised by Prof. Xun Cao and Prof. Yang Yu. He received the B.S. degree from Beijing Institute of Technology in 2021. His research interests include computational photography and implicit neural representation.
\end{IEEEbiography}

\begin{IEEEbiography}[{\includegraphics[width=1in,height=1.25in,clip,keepaspectratio]{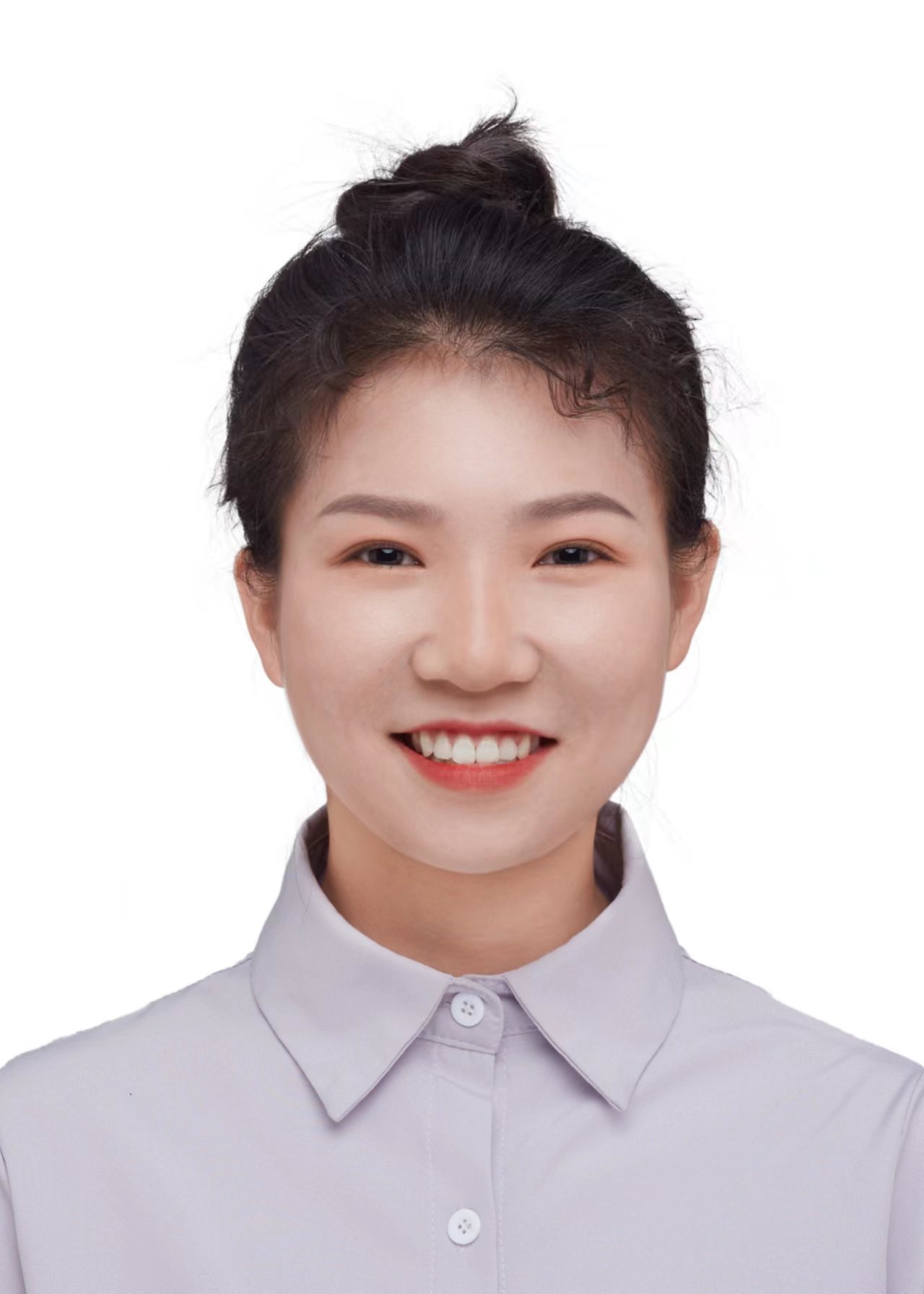}}]{Fengyi Liu} is a graduate student at the School of Electronic Science and Engineering, Nanjing University.  She received the B.S. degrees from Sichuan University in 2022. Her research interests include novel view synthesis and implicit neural representations.
\end{IEEEbiography}

\begin{IEEEbiography}[{\includegraphics[width=1in,height=1.25in,clip,keepaspectratio]{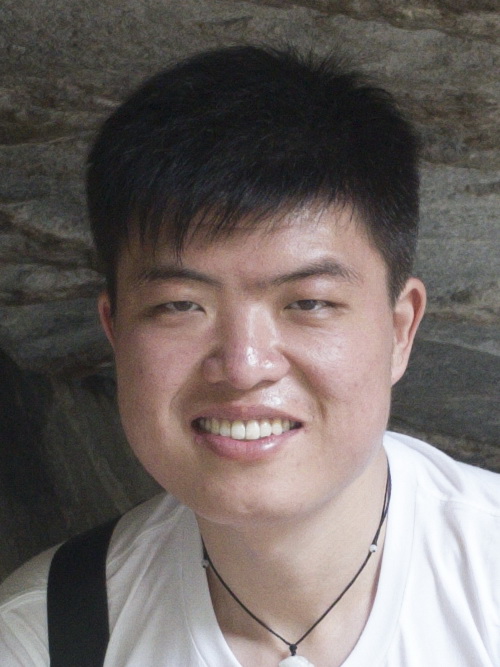}}]{Qi Zhang} is currently a researcher with Tencent AI Lab. He received the Ph.D. degree from the School of Computer Science at Northwestern Polytechnical University in 2021. His research interests include 3D vision reconstruction, light field imaging and processing, multi-view geometry and application.
\end{IEEEbiography}

\begin{IEEEbiography}[{\includegraphics[width=1in,height=1.25in,clip,keepaspectratio]{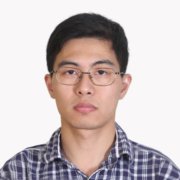}}]{You Zhou} is an associate researcher at the School of Electronic Science and Engineering, Nanjing University, China. He received his Ph.D. degree (January 2019) from Department of Automation at Tsinghua University and his B.S. degree (July 2012) from Department of Communication Engineering at East China Normal University, respectively. He was an exchange researcher at Biomedical Engineering Department, University of Connecticut from December 2016 to May 2017. His research focuses on computational microscopy and computational optics. He is a member of IEEE.
\end{IEEEbiography}

\begin{IEEEbiography}[{\includegraphics[width=1in,height=1.25in,clip,keepaspectratio]{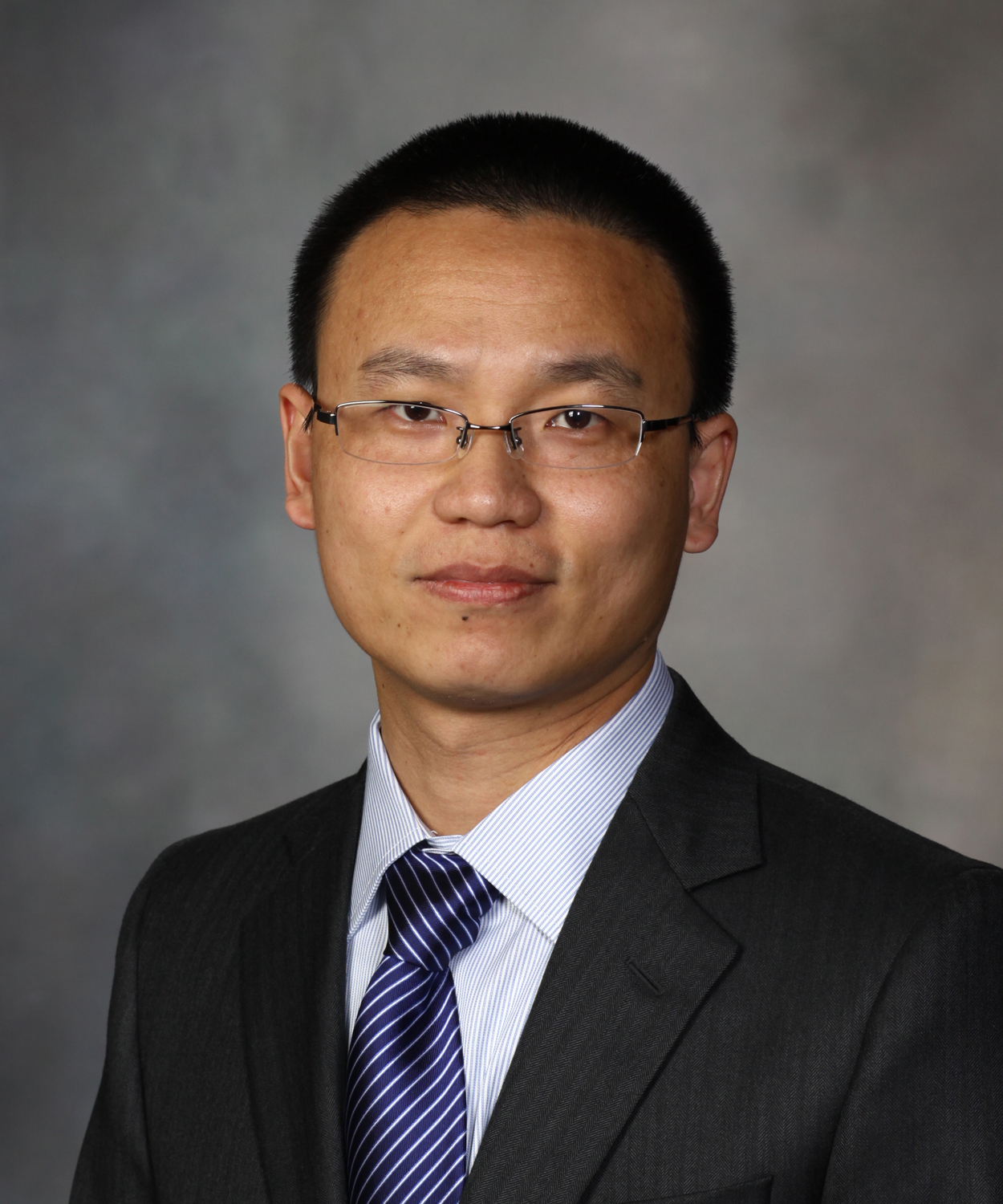}}]{Yi Lin}
Dr. Yi Lin is a cardiovascular surgeon and associate professor at Zhongshan Hospital of Fudan University. He received medical education at Shanghai Medical School of Fudan University and then completed his Surgery, Cardiovascular and Thoracic Surgery residency at Zhongshan Hospital of Fudan University. After the clinical training, he spent one year as surgical research fellow at Mayo Clinic in Rochester followed by advanced cardiovascular surgery training at Mayo Clinic in Rochester. He returned to Zhongshan Hospital of Fudan University to join the staff in 2016 where he is currently an Attending Surgeon and Associate Professor of Cardiovascular Diseases. His primary research interest includes innovative medical imaging techniques and robot-assisted therapies in cardiovascular diseases. 
\end{IEEEbiography}

\begin{IEEEbiography}[{\includegraphics[width=1in,height=1.25in,clip,keepaspectratio]{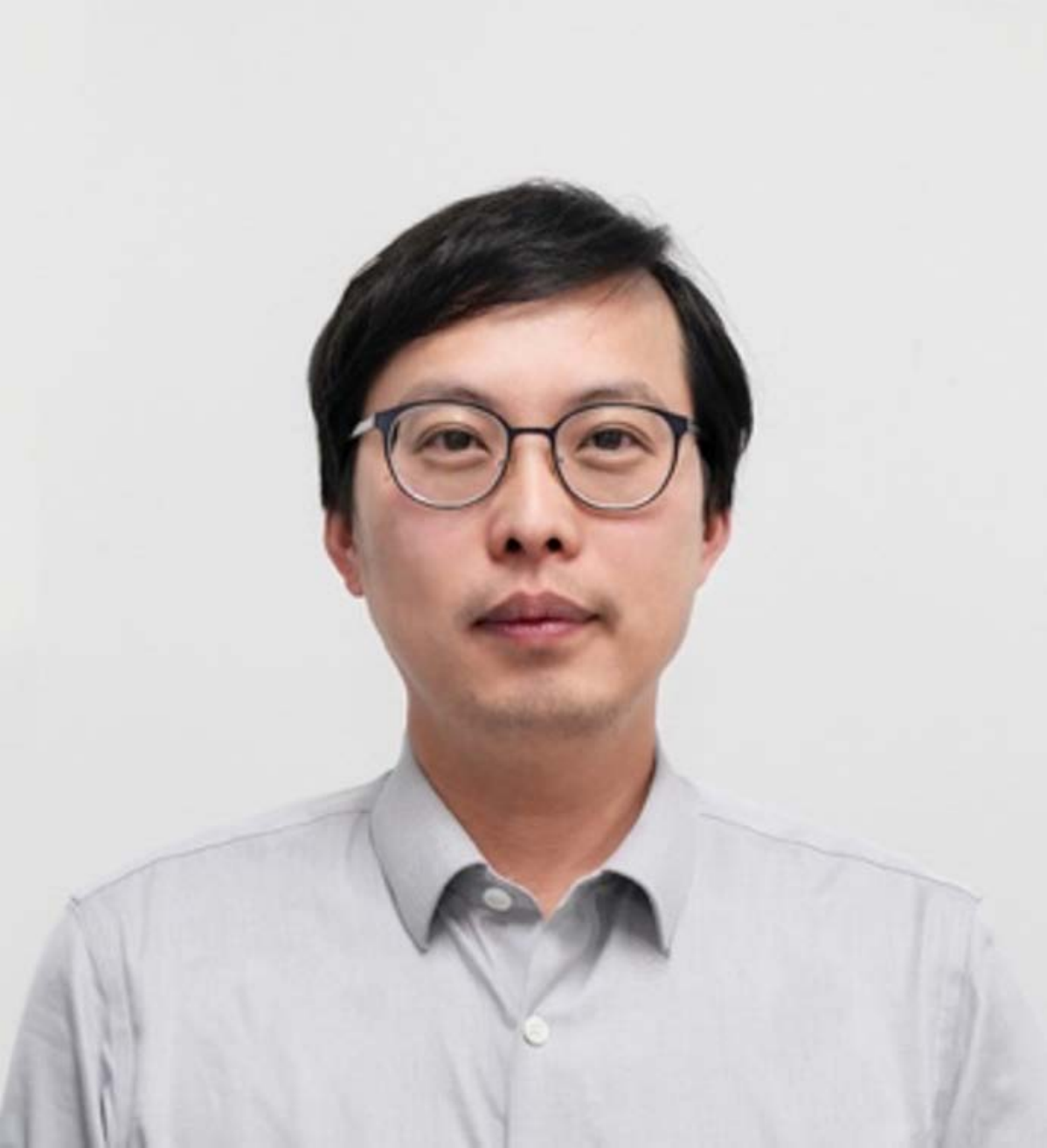}}]{Zhan Ma} (SM'19) is now on the faculty of Electronic Science and Engineering School, Nanjing University, Nanjing, Jiangsu, 210093, China. He received the B.S. and M.S. degrees from the Huazhong University of Science and Technology, Wuhan, China, in 2004 and 2006 respectively, and the Ph.D. degree from the New York University, New York, in 2011. From 2011 to 2014, he has been with Samsung Research America, Dallas TX, and  Futurewei Technologies, Inc., Santa Clara, CA, respectively. His research focuses on the learning-based video coding, and smart cameras. He is a co-recipient of the 2018 PCM Best Paper Finalist, 2019 IEEE Broadcast Technology Society Best Paper Award, 2020 IEEE MMSP Image Compression Grand Challenge Best Performing Solution, and 2023 IEEE WACV Best Algorithm Paper Award.
\end{IEEEbiography}

\begin{IEEEbiography}[{\includegraphics[width=1in,height=1.25in,clip,keepaspectratio]{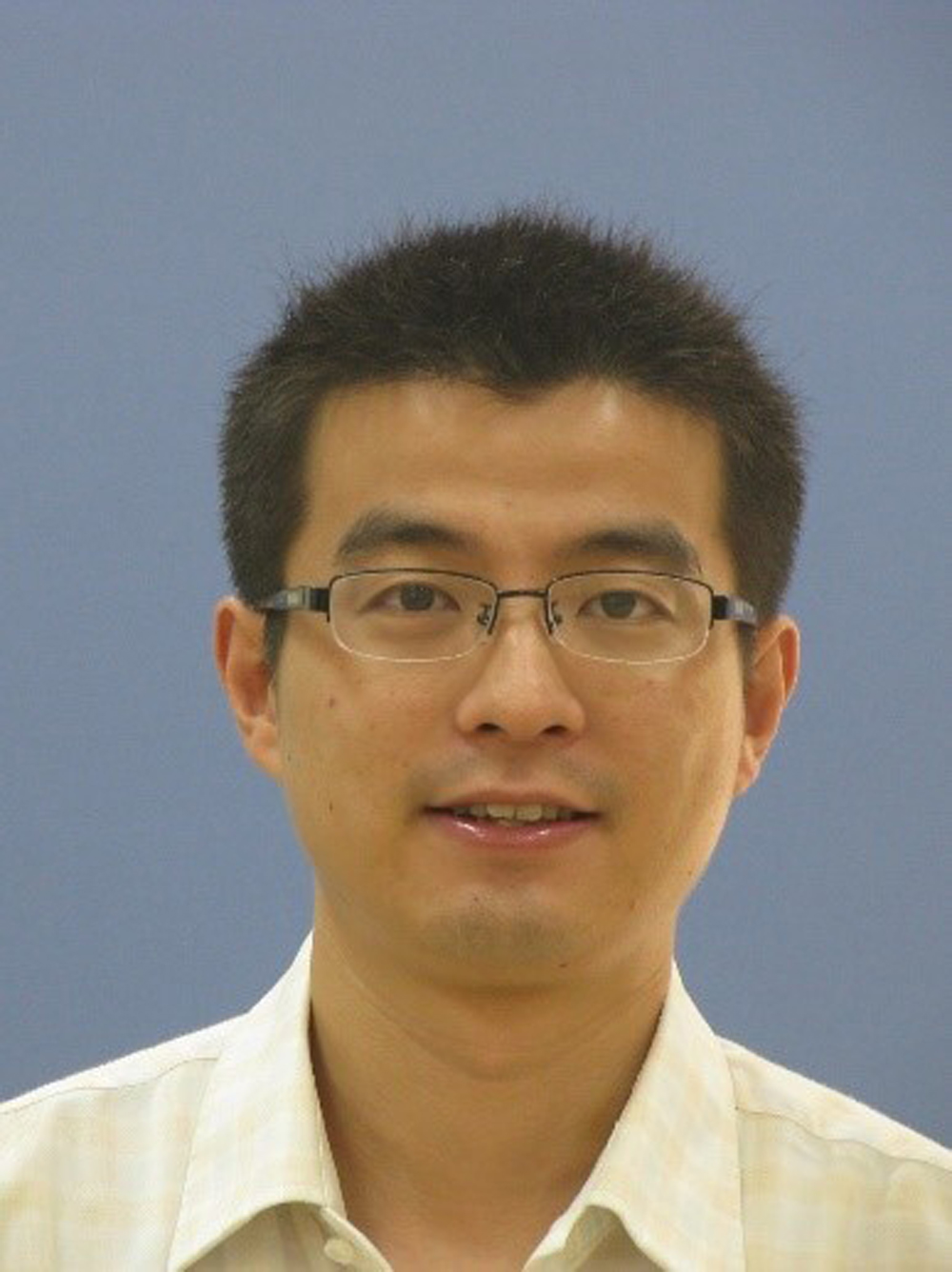}}]{Xun Cao} received the B.S. degree from Nanjing University, Nanjing, China, in 2006, and the Ph.D. degree from the Department of Automation, Tsinghua University, Beijing, China, in 2012. He held visiting positions with Philips Research, Aachen, Germany, in 2008, and Microsoft Research Asia, Beijing, from 2009 to 2010. He was a Visiting Scholar with the University of Texas at Austin, Austin, TX, USA, from 2010 to 2011. He is currently a Professor with the School of Electronic Science and Engineering, Nanjing University. His current research interests include computational photography and image-based modeling and rendering.
\end{IEEEbiography}


%\newpage



%\vfill


\end{document}